\documentclass{article}

\PassOptionsToPackage{numbers}{natbib}
\usepackage[preprint]{neurips_2026}

\usepackage[utf8]{inputenc} 
\usepackage[T1]{fontenc}    
\usepackage{hyperref}       
\usepackage{url}            
\usepackage{booktabs}       
\usepackage{amsfonts}       
\usepackage{nicefrac}       
\usepackage{microtype}      

\usepackage{titlesec}

\usepackage{titlesec}
\titlespacing*{\paragraph}{0pt}{-1mm}{.2em}

\usepackage{color}	
\usepackage{xspace}
\usepackage[table, dvipsnames]{xcolor}
\usepackage{colortbl}
\usepackage{soul}

\usepackage{graphicx}
\usepackage{wrapfig}
\usepackage{color}
\usepackage{enumerate}
\usepackage{hyperref}
\usepackage[font={small,sf}]{caption}
\usepackage{amsthm}
\usepackage{floatflt}
\usepackage{lipsum}
\usepackage{multirow}
\usepackage{soul}
\sethlcolor{black!10}
\usepackage{mdframed}
\usepackage{amsmath,amsfonts}
\usepackage{comment}
\usepackage{subcaption}
\usepackage{amssymb}
\usepackage{enumitem}
\usepackage{csquotes}
\usepackage{tabularx}
\usepackage{tabularx}
\usepackage{enumitem}
\usepackage{tikz}
\usetikzlibrary{positioning}
\usepackage{mathtools}
\usepackage[a-1b]{pdfx}  
\usepackage[T1]{fontenc}
\usepackage{microtype}
\usepackage{bbm}
\usepackage{fnpct}  
\usepackage{tcolorbox}

\graphicspath{{./figures/}}

\usepackage{algorithm}
\usepackage[noend]{algpseudocode}

\DeclarePairedDelimiterX\set[1]\lbrace\rbrace{\,#1\,}

\usepackage{tikz}
\usetikzlibrary{positioning}

\DeclareMathOperator*{\argmax}{arg\,max}

\newcommand{\stitle}[1]{\vspace{1mm}\noindent{\textbf{#1}}.}

\newcommand{\at}[1]{{\tt \small #1}\xspace}

\renewcommand{\Re}{\mathbb{R}}%

\newtheorem{definition}{Definition}
\newtheorem{problem}{Problem}
\newtheorem{example}{Example}

\newcommand{\point}{p}

\newtcolorbox{highlightbox}{
  colback=blue!10,
  colframe=blue!20,
  arc=0.5mm, 
  fonttitle=\bfseries,
  boxrule=0mm,
  boxsep=0mm,
  left=0mm,
  right=0mm,
  top=0mm,
  bottom=0mm
}

\newtcolorbox{examplebox}{
  colback=blue!10,
  colframe=blue!20,
  arc=2mm, 
  fonttitle=\bfseries,
  boxrule=0mm,
  boxsep=1mm,
  left=0mm,
  right=0mm,
  top=0mm,
  bottom=0mm
}

\newtcolorbox{defbox}{
  colback=orange!10,
  colframe=orange!20,
  arc=2mm, 
  fonttitle=\bfseries,
  boxrule=0mm,
  boxsep=1mm,
  left=0mm,
  right=0mm,
  top=0mm,
  bottom=0mm
}

\newtcolorbox{pbox}{
  colback=black!5,
  colframe=black!30,
  arc=2mm, 
  fonttitle=\bfseries,
  boxrule=0mm,
  boxsep=1mm,
  left=0mm,
  right=0mm,
  top=0mm,
  bottom=0mm
}

\usepackage{tcolorbox}
\tcbuselibrary{skins,breakable}

\newtcolorbox{wbox}[1]{enhanced, colback=white, colframe=red!65!black,
  colbacktitle=red!65!black, coltitle=white, fonttitle=\bfseries\scriptsize,
  title=#1, left=3pt, right=3pt, top=1pt, bottom=1pt, boxrule=0.5pt,
  arc=1.6pt, sharp corners=southwest, fontupper=\scriptsize}
\newtcolorbox{nbox}[1]{enhanced, colback=white, colframe=teal!75!black,
  colbacktitle=teal!75!black, coltitle=white, fonttitle=\bfseries\scriptsize,
  title=#1, left=3pt, right=3pt, top=1pt, bottom=1pt, boxrule=0.5pt,
  arc=1.6pt, sharp corners=southwest, fontupper=\scriptsize}

\title{Sparse Attention as a Range Searching Problem: Towards an Inference-Efficient Index for KV Cache}

\author{%
  Mohsen Dehghankar \\
  Department of Computer Science\\
  University of Illinois Chicago\\
  \texttt{mdehgh2@uic.edu} \\
  \And
  Abolfazl Asudeh \\
  Department of Computer Science \\
  University of Illinois Chicago \\
  \texttt{asudeh@uic.edu}
}

\begin{document}

\maketitle

\begin{abstract}
    Sparse attention improves LLM inference efficiency by selecting a subset of key–value (KV) entries, but at the cost of potential accuracy degradation. In particular, omitting critical KV entries can induce substantial errors in model outputs. Existing methods typically operate under fixed or adaptive token budgets and provide empirical robustness or partial theoretical guarantees, yet they do not ensure zero false negatives across decoding steps, particularly since the set of relevant tokens is both query- and step-dependent.
    Our empirical observations confirm that missing even one critical key can lead to sharp error spikes, especially in long-output reasoning tasks where the set of important tokens varies throughout decoding. This observation motivates the need for indexing methods that dynamically adapt to these variations across decoding steps while guaranteeing a full recall of the relevant keys above a certain threshold.
    We address this challenge by reformulating sparse attention as the computational geometry problem of halfspace range searching. However, existing range searching data structures are not suitable for modern LLM inference due to their computational and implementation overheads. To overcome this, we introduce {\bf Louver}, a novel index structure tailored for efficient KV cache retrieval. {\bf Louver} (i) guarantees zero false negatives with respect to a specified threshold in both theory and practice, (ii) is lightweight to integrate into existing LLM inference pipelines, and (iii) incorporates hardware-aware optimizations for both CPU and GPU executions.
    Our experiments demonstrate that {\bf Louver} outperforms prior sparse attention methods in both accuracy and runtime, and is faster than highly optimized dense implementations such as FlashAttention. These results highlight that recall guarantees are a critical and overlooked dimension of sparse attention, and open a new direction for building theoretically grounded, efficient KV cache indices. Code is available at \url{https://github.com/UIC-InDeXLab/Louver}.

\end{abstract}

\vspace{-3mm}
\section{Introduction}

\vspace{-4mm}
\begin{flushright}
\small{\textit{``... the ocean would be less because of that missing drop.''}
 ---Mother Teresa}
\end{flushright}
\vspace{-2mm}

Large Language Models (LLMs) rely on the Key-Value (KV) cache to reuse previously computed key and value vectors for efficient inference. While effective, the KV cache introduces significant memory and computational overhead, as each newly generated token attends to all prior tokens. A key observation is that attention is inherently sparse: for any query token, only a small subset of past tokens, called ``active tokens'' meaningfully contributes to its representation~\cite{sparsetransformer, longformer}. This sparsity has motivated a large body of work on accelerating inference via selective attention over the KV cache.

 \begin{examplebox}
\begin{example}\label{ex-1} \small{
As a running example, let the last token in a context window be $t_n=$\at{[paper]}, with the vector $x_n$ encoding the ``description'' of the general concept \at{[paper]}.
Suppose the context contains:

\begin{quote}
    \at{[...We attended \textcolor{red}{NeurIPS}, \textcolor{red}{KDD}, \textcolor{red}{SIGIR}, and \textcolor{red}{VLDB} last year, to present our \textcolor{olive}{papers}]}
\end{quote}
\vspace{.5mm}
    The highlighted tokens \at{[NeurIPS]}, \at{[KDD]}, \at{[SIGIR]}, and \at{[VLDB]}
    are the {\em active tokens} that contribute to refining $x_n$, shifting it toward concepts such as Computer Science, Machine Learning, Information Retrieval, Data Mining, Data Management, etc, by inducing an update $\tilde{o}_n$ as $x_n\gets x_n+\tilde{o}_n$.
    Importantly, 
    \begin{enumerate}
        \item the number of active tokens is not fixed, as it depends on the context, and
        \item each token may contribute a distinct semantic aspects, {making their inclusion critical}.
    \end{enumerate}
    }
\end{example}
\end{examplebox}

\vspace{-1mm}
Existing KV-cache acceleration techniques often rely on selecting a fixed number $k$ of tokens per query (e.g., top-$k$ retrieval or approximate nearest neighbors)~\cite{retrievalattention, magicpig, pqcache, hashattention, squeezedattention}. Conversely, the number of active tokens is inherently {\em query-dependent} and cannot be determined a priori. As illustrated in Example~\ref{ex-1}, the number of relevant tokens (e.g., conferences attended) can vary arbitrarily, making any fixed value of $k$ either insufficient or wasteful.
\\
However, even more critically, these methods can introduce {\bf false negatives}, i.e., they may miss tokens that significantly contribute to the attention computation. This issue arises from two sources: (i) the approximate nature of the underlying retrieval algorithms, and (ii) the use of a fixed $k$, which enforces an artificial cap on the number of retrieved tokens. 
Missing an active token can lead to incomplete or biased updates of the query embedding. In Example~\ref{ex-1}, omitting \at{[VLDB]} would remove the data management aspect from $\tilde{o}_n$, degrading the semantic fidelity of the model's representation.
As a result, unlike many retrieval settings, {\em sparse attention is highly sensitive to missing relevant tokens}. This observation fundamentally challenges the prevailing design of {\em approximate} or top-$k$ retrieval for KV caching and calls for mechanisms that prevent {\em false negatives}.


\vspace{-2mm}
\stitle{Our Perspective}
In this paper, we revisit KV-cache sparsification through the lens of \underline{\em exact} active-token retrieval. 
We argue that the problem should not be formulated as selecting a fixed-size or as an approximation objective, but rather as identifying all relevant tokens whose contribution exceeds a query-dependent threshold.
In particular, our contributions can be summarized as follows:
\vspace{-1mm}
\begin{itemize}[leftmargin=*, itemsep=0pt, topsep=0pt]
    \item 
    We reduce the sparse attention problem to the {\em range searching problem}. Through this reduction, we show that an exact solution for the range searching problem retrieves all tokens whose similarity with the query exceeds a dynamically determined threshold. This formulation addresses both limitations: it eliminates the need for a fixed budget $k$ and prevents false negatives. 
    \item
    The existing range-searching data structures are not suitable for LLM inference. 
    We identify a set of index design challenges for KV-cache sparsification.
    \item
    Building on the design challenges, we develop an efficient indexing framework, called {\sc Louver}, that supports dynamic, threshold-based retrieval over the KV cache, ensuring exactness while achieving significant computational savings.
    \item 
    We introduce various system-level optimizations to make Louver deployment-ready to integrate into existing LLM-inference pipelines, 
    develop a lightweight oracle for dynamically identifying the relevance threshold, and incorporate hardware optimizations for CPU and GPU executions. 
    \item We conduct comprehensive experiments across fixed-budget, adaptive-budget, KV-offloading, and dense-attention baselines, showing that \textsc{Louver} consistently achieves higher accuracy and lower latency than existing methods, while matching or exceeding dense-attention accuracy and achieving up to {\bf 15.3$\times$ GPU} and {\bf 10.3$\times$ CPU} speedups at 40k context length.
\end{itemize}

\paragraph{Related Work.}
Prior work on sparse attention for long-context inference spans eviction-based methods, which discard tokens permanently~\cite{streamingllm,snapkv,h2o}, and retrieval-based methods, which dynamically retrieve relevant KV entries using ANN- or hashing-based indexing~\cite{retrievalattention,magicpig,hashattention}. Existing approaches typically rely on approximate retrieval or fixed/adaptive token budgets~\cite{twilight,pyramidkv,blasst}, without guaranteeing retrieval of all keys whose attention contribution exceeds the relevance threshold. In contrast, our work formulates sparse attention as an exact range-search problem, enabling dynamically updated indexing with adaptive per-query thresholds. Our approach is complementary to orthogonal acceleration techniques such as FlashAttention~\cite{flashattention2} and vLLM~\cite{vllm}. Please refer to Appendix~\ref{app:related} for the detailed discussion of the related work.

\vspace{-5mm}
\section{Preliminaries}
\label{sec:prelim}
\vspace{-4mm}

\paragraph{Attention.}
Let $X \in \mathbb{R}^{n \times d}$ denote a sequence of $n$ token
representations, where $d$ is the model's embedding dimension.
In \emph{multi-head attention} (MHA), $H$ attention heads operate in
parallel, each with its own learned projection matrices, where
$d_h = d/H$ is the \emph{head dimension}.
Within each head, $X$ is projected into queries, keys, and values as
$Q = XW_Q$, $K = XW_K$, $V = XW_V$ with
$W_Q, W_K, W_V \in \mathbb{R}^{d \times d_h}$ and
$Q, K, V \in \mathbb{R}^{n \times d_h}$.

For query $q_i \in \mathbb{R}^{d_h}$ and key $k_j \in \mathbb{R}^{d_h}$,
the \emph{attention score} is $s_{ij} = \langle q_i, k_j\rangle / \sqrt{d_h}$.
Scores are normalized via softmax to obtain \emph{attention weights}: $a_{tj} = \operatorname{softmax}_j\bigl(s_{t1}, \dots, s_{t,t-1}\bigr).$
The per-head output $o_i = \sum_{j \le i} a_{ij} v_j \in \mathbb{R}^{d_h}$
is computed for each head, concatenated across all $H$ heads, and
projected back to $\mathbb{R}^d$ to give $\tilde{o}_i \in \mathbb{R}^d$, which is then
added to the residual stream as $x_i \leftarrow x_i + \tilde{o}_i$
before being passed to the next layer. We focus on a single attention head throughout the paper
and use $d$ in place of $d_h$, slightly abusing notation.

\paragraph{Prefilling and Decoding.}
Transformer inference proceeds in two phases.
In \emph{prefilling}, the prompt $x_{1:n}$ is processed in a single
parallel forward pass, computing $K, V \in \mathbb{R}^{n \times d}$
simultaneously, producing the initial set of key and value vectors
$\{(k_j, v_j)\}_{j=1}^n$ that serve as context for generation.
Since all tokens are available at once, prefilling is parallelizable
and computationally straightforward.
\\
In \emph{decoding}, tokens are generated one at a time autoregressively:
at step $t$, the model computes query $q_t \in \mathbb{R}^{d}$ and
attends to \emph{all} $t-1$ previously computed keys and values,
  $o_t \;=\; \sum_{j=1}^{t-1} a_{tj}\, v_j,$
where $a_{tj}$ is the attention weight restricted to the $t-1$ cached keys.
The output $o_t$ is added to the residual stream and passed to the next layer; after the full forward pass, the final hidden state is then projected onto the
vocabulary to sample the next token.
\\
Unlike prefilling, decoding is inherently sequential, and its cost grows
with each step, making it the dominant computational bottleneck.
This is especially observed in long-output reasoning tasks, where the model generates thousands of tokens~\cite{surveyefficientattention, surveykvmanagement}. While our methods extend naturally to prefilling, this paper focuses primarily on the decoding phase.

\paragraph{KV Cache.}
To avoid recomputing keys and values from scratch at each decoding step,
the model stores all past key--value pairs in a \emph{KV cache}:
\(
  \mathcal{KV}_t \;=\; \bigl\{(k_j,\, v_j)\bigr\}_{j=1}^{t-1},
  \qquad k_j, v_j \in \mathbb{R}^{d}.
\)
\\
At step $t$, attention requires computing $t-1$ inner products
$\langle q_t, k_j \rangle$ and a weighted sum over all $t-1$ cached
values, so the cost per step grows linearly with sequence length.
Grouped query attention (GQA)~\cite{gqa} further reduces the memory footprint of the KV cache by
sharing a single key--value head among a group of $G$ query heads. Standard MHA corresponds to $G = 1$.


\paragraph{Sparse Attention.}
Sparse attention aims to reduce the cost of attention by computing it over only a subset of the KV cache. Instead of attending to all $t-1$ past keys at decoding step $t$, the goal is to identify a subset of \emph{relevant} keys (and their corresponding values) that contribute most to the output.
Specifically, given a query $q_t$, sparse attention retrieves a subset of keys
$\mathcal{K}^* \subseteq \{k_1, \dots, k_{t-1}\}$ and computes $o_t \;=\; \sum_{k_j \in \mathcal{K}^*} a_{tj}\, v_j.$ 
\\
Aligned with prior work~\cite{twilight, tactic, sampleattention}, given a threshold $\tau$, we define the sparse attention problem as finding the set of relevant keys satisfying $\langle q_t, k_j \rangle \ge \tau$; 
Formally,

\vspace{-2.5mm}
\begin{defbox}
\begin{problem}\label{problem:sparseattention}
    Given a query vector $q_t\in \mathbb{R}^d$ at decoding step $t$, and a set of key vectors $\mathcal{K} = \{k_j\in \mathbb{R}^d\}_{j=1}^{t-1}$, 
    find the subset $\mathcal{K}^* \subseteq \mathcal{K}$, where
    \(
        \mathcal{K}^* \;=\; \bigl\{k \in \mathcal{K} \mid \langle q_t, k \rangle \geq \tau \bigr\}.
    \)
\end{problem}
\end{defbox}

\vspace{-6mm}
\subsection{Problem Motivation based on Experimental Observations}
\vspace{-3mm}

Before discussing our approach to solving the problem and our reduction to range searching, in the following, we provide two observations that further motivate (a) formulating sparse attention as Problem~\ref{problem:sparseattention} and (ii) solving it {exactly} and not approximately.

\begin{figure*}[t]
    \centering
    \begin{subfigure}[t]{0.25\linewidth}
        \centering
        \includegraphics[width=\linewidth]{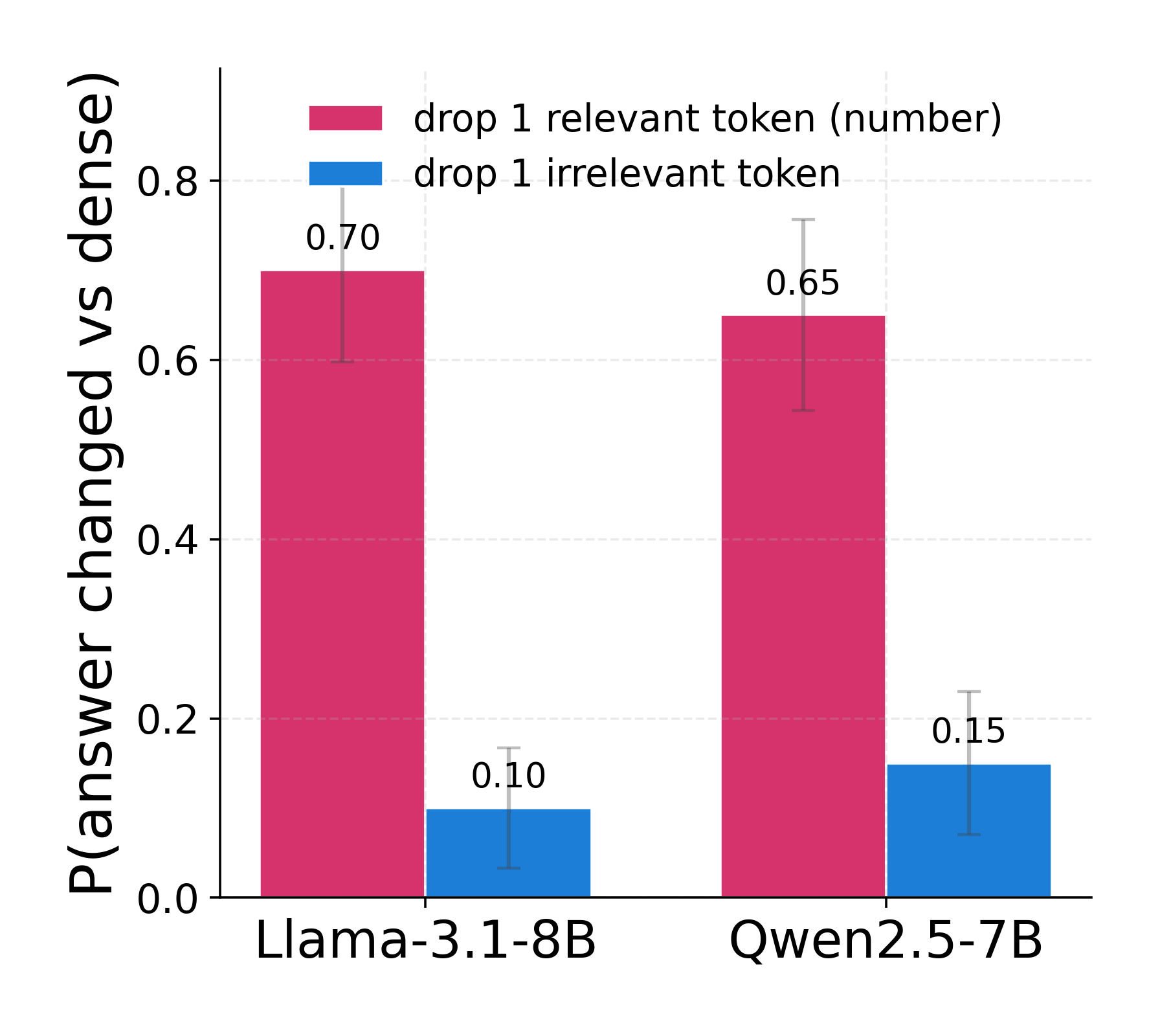}
        \vspace{-8mm}
        \caption{}  
        \label{fig:drop-error}
    \end{subfigure}
    \hfill
    \begin{subfigure}[t]{0.36\linewidth}
        \centering
        \includegraphics[width=\linewidth]{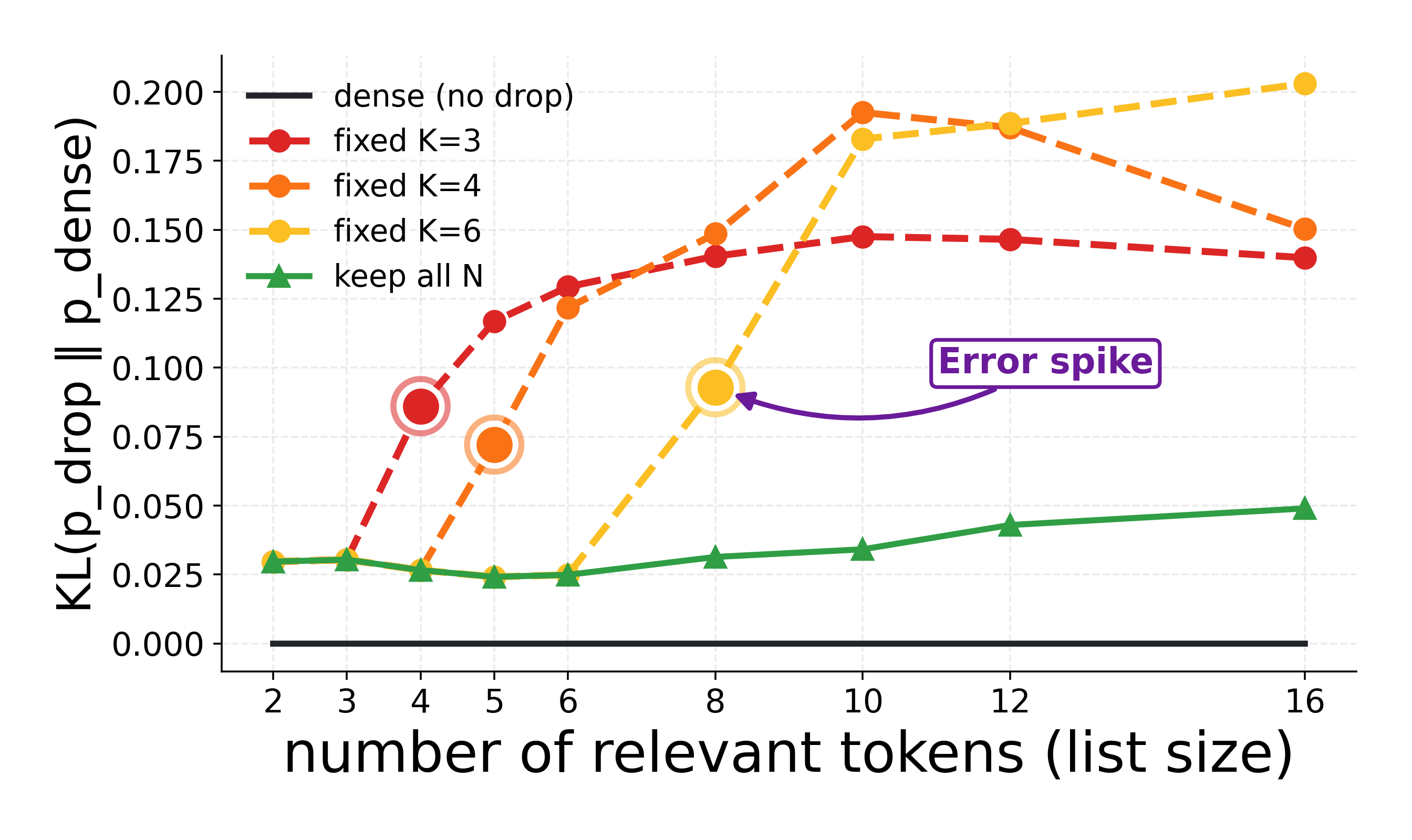}
        \vspace{-8mm}
        \caption{}  
        \label{fig:error-spike}
    \end{subfigure}
    \begin{subfigure}[t]{0.36\linewidth}
        \centering
        \includegraphics[width=\linewidth]{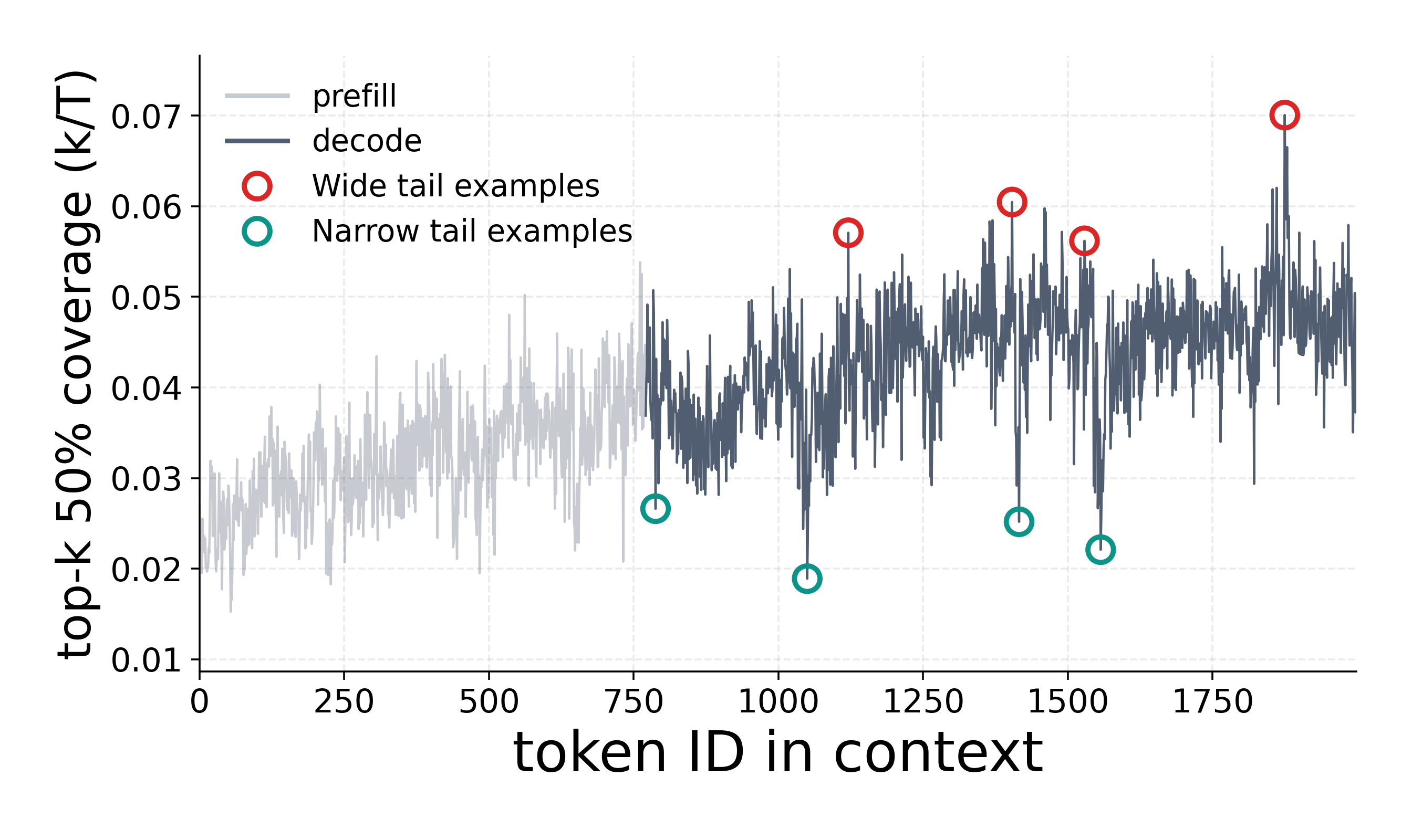}
        \vspace{-8mm}
        \caption{}  
        \label{fig:decoding}
    \end{subfigure}
    \vspace{-2mm}
    \caption{\textbf{(a)} Effect of dropping tokens. We remove either (i) one relevant token (a randomly selected number in the list) or (ii) one randomly selected non-relevant token. Dropping a relevant token significantly increases the error.
    \textbf{(b)} Error spike in fixed-$K$ sparse attention as the list size $N$ increases. The green line corresponds to retaining all relevant tokens (no false negatives). \textbf{(c)} Top-$k$ coverage across decoding steps, showing significant oscillation and indicating that the number of relevant tokens varies per step (see Observation~2). See Appendices~\ref{app:false-negativity} and \ref{app:score-distribution} for details.}
    \label{fig:main}
    \vspace{-6mm}
\end{figure*}

\paragraph{Observation 1 (Sensitivity to False Negatives).}
Omitting relevant context tokens leads to significant degradation in attention quality. While an illustrative example is provided in the Introduction, we further demonstrate this effect through a simple experiment.
We consider the prompt: 
\textit{``Given a list of numbers $[x_1, x_2, \ldots, x_N]$, compute the sum of all elements. Answer with an integer.''}
We vary the list size $N$ and simulate sparse attention by dropping KV-cache entries. 
In particular, we remove (i) one token corresponding to a number in the list (i.e., a relevant token), and (ii) one randomly selected non-relevant token.
Figure~\ref{fig:drop-error} shows that removing a relevant token results in a significant increase in error compared to removing non-relevant tokens. This confirms that all elements in the list are necessary for correctly generating the answer, and that omitting any of them (i.e., introducing false negatives) substantially degrades performance.
We further evaluate fixed-$K$ sparse attention under varying list sizes. As shown in Figure~\ref{fig:error-spike}, fixed-budget methods exhibit sharp error increases as $N$ grows. This is expected, since the number of relevant tokens grows with the input size while the attention budget remains fixed. In practice, the number of important tokens is unknown, and dropping them can lead to such error spikes. In contrast, retaining all relevant tokens eliminates this issue.
As a result, achieving low error (i.e., avoiding error spikes) requires retaining {\bf all relevant} KV-cache entries, whose number is not fixed in practice.
Prior work has also identified the necessity of adaptive sparsity beyond fixed K~\cite{twilight, tactic, sampleattention}, typically using coverage-based strategies that select tokens to meet a score threshold. However, developing approximate solutions, these approaches can still omit relevant tokens, as highlighted in the Appendix~\ref{app:false-negativity}.

\paragraph{Observation 2 (Context- and Query-Dependent Score Distribution).}
Our second observation takes a more computational perspective, focusing on attention scores, i.e., the dot product between query and key vectors. We analyze attention score distributions across different decoding steps in a long reasoning output.
For each decoding step, we quantify the tail of the distribution using \emph{top-$k$ coverage}. Larger values indicate that attention is spread over more tokens (wider tails), while smaller values indicate more concentrated attention\footnote{More details in Appendix~\ref{app:score-distribution}.}.
As shown in Figure~\ref{fig:decoding}, this metric exhibits strong oscillation across decoding steps. In particular, during reasoning phases such as transitions between thoughts, backtracking, or planning next steps, attention tends to be broader, involving many relevant tokens. In contrast, during more localized computations (e.g., intermediate calculations), attention becomes more concentrated. Explicit examples of these regions are provided in Appendix~\ref{app:score-distribution}.

Observations~1 and~2 indicate that the attention score distribution is inherently \emph{context- and query-dependent}, varying significantly across decoding steps. As a result, it cannot be well-approximated by a fixed pattern or budget throughout inference.
Prior work has also observed this dynamic behavior across different attention heads and layers~\cite{pyramidinfer, pyramidkv, adakv}. Together, these findings suggest that attention distributions are dynamic both horizontally (across tokens) and vertically (across layers), motivating the formulation of Problem~\ref{problem:sparseattention} and the need for exact solutions to it.


\vspace{-4mm}
\section{Reducing Sparse Attention to Halfspace Range Searching}
\label{sec:sparse-as-range-searching}
\vspace{-3mm}

In order to solve Problem~\ref{problem:sparseattention}, we reduce it to the computational geometry problem of {\em range searching}.

Range searching is defined on a {\em range space} $(P,\mathcal{R})$ with a finite set of points $P = \{p_1, \cdots, p_m\}$, where each point $\point_i$ lies in a $d$-dimensional space $\Re^d$~\cite{har2011geometric}. Associated with $P$ is a family of subsets $\mathcal{R}$, referred to as ranges (e.g., rectangular boxes). Given a range $R\in\mathcal{R}$, the range searching problem retrieves the subset of $P$ that falls inside $R$. Formally,

\vspace{-2.5mm}
\begin{defbox}
\begin{problem}[Range Searching~\cite{rangesearching}]\label{problem:rangesearching}
    Given a range space $(P,\mathcal{R})$, preprocesses $P$ into an index so that for any range $R\in\mathcal{R}$, one can retrieve $P \cap R \;=\; \{p \in P \,\vert\, p \in R\}$, in time sublinear in $|P|$.
\end{problem}
\end{defbox}

A critical requirement in the range searching problem---and one we enforce throughout this paper---is
\emph{100\% recall}: all points in $P \cap R$ must be returned, with no false negatives.

Having reviewed the range searching problem, we next provide our reduction:

\paragraph{Reduction.}
Given an instance of the sparse attention problem (Problem~\ref{problem:sparseattention}) 
with the query vector $q_t\in \mathbb{R}^d$ and a set of key vectors $\mathcal{K} = \{k_j\in \mathbb{R}^d\}_{j=1}^{t-1}$ in the KV cache at decoding step $t$, 
define the following range searching problem (Problem~\ref{problem:rangesearching}):
\begin{enumerate}
    \item {\em Points:} For each key $k_j\in\mathcal{K}$, we add a point\footnote{From now on, we use terms `points' and `keys' interchangeably.} to the point set. That is $P\gets \mathcal{K}$.
    \item {\em Ranges:} For the pair $(q_t,\tau)$, we define a {\em halfspace} $R$. A halfspace in $\mathbb{R}^d$ is the set of points on one side of a hyperplane. Specifically, we define the halfspace $R$ as
    \small{\[R\gets \mathrm{H}(q_t,\tau) \;=\; \bigl\{p \in \mathbb{R}^d \,\vert\, \langle q_t,\, p\rangle \ge \tau\bigr\}.\]}
    \item {\em Output mapping:} The output of the range searching problem is $P\cap R = \{k\in\mathcal{K}\,\vert\, \langle q_t,\, k\rangle \ge \tau\}$. As a result, $\mathcal{K}^*\gets P\cap R$.
\end{enumerate}

\begin{wrapfigure}{r}{0.33\linewidth}
    \centering
    \vspace{-9mm}
    \includegraphics[width=\linewidth]{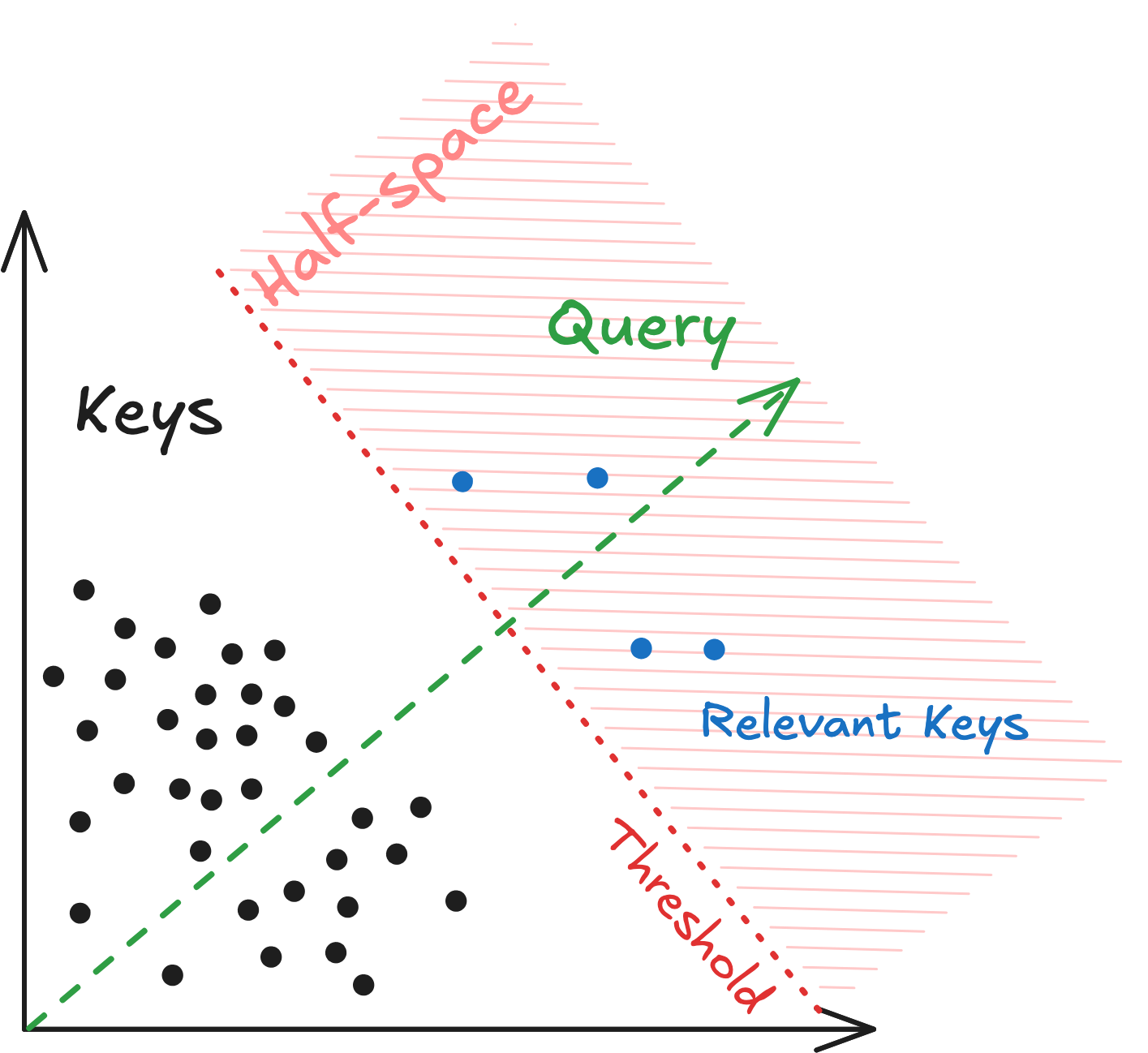}
    \vspace{-6mm}
    \caption{Illustrating the reduction of sparse attention to halfspace range searching (Example~\ref{ex-2}).}
    \label{fig:reduction}
    \vspace{-6mm}
\end{wrapfigure}

\begin{examplebox}
\begin{example}\label{ex-2} \small{
To better illustrate the reduction, let us consider the toy example presented in Figure~\ref{fig:reduction}.
In this example, $d=2$, i.e., $q_t, k_j\in\mathbb{R}^2$.
Each key $k_j\in\mathcal{K}$ is shown as a point in the figure.
The query $q_t$ is translated to the green direction, and the threshold $\tau$ specifies the dashed line orthogonal to it. The halfspace is the dashed open region bounded by the threshold line.
The relevant keys are the blue points inside the halfspace.
}
\end{example}
\end{examplebox}
Reducing sparse attention to halfspace range searching enables building an index over the keys in the KV cache that supports halfspace range searching queries. A solution to halfspace range searching satisfies
{\bf Observation~1}, since by the 1-1 reduction mapping, all relevant keys fall inside the range.
It also naturally addresses {\bf Observation~2}: given a threshold $\tau$, the index returns all keys that exceed it, with no fixed assumptions on the number or identity of retrieved keys.
As a result, the retrieved set varies per decoding step, adapting dynamically to the query and the current context. 

\paragraph{Remark (Comparison to Similarity Search).}
Traditionally, the sparse attention problem is reduced to a similarity search problem~\cite{retrievalattention, magicpig, pqcache, hashattention, squeezedattention}, often adapting techniques and indices used for {approximate} nearest neighbor (ANN) search and maximum inner product
search (MIPS). While range searching and similarity search are related problems, they are fundamentally different and suited to different scenarios~\cite{simplexrangesearching}.
The first difference is in the objectives: ANN (resp. MIPS) indices are built to {\em approximately} find the {\em top-$k$} nearest neighbors (resp. max inner product).
The second key difference concerns {\em normalization}.
To apply similarity search indexes to KV cache retrieval, one must adopt cosine similarity or angular distance as the retrieval metric~\cite{retrievalattention, hashattention, magicpig}.
Similarity search indexes based on cosine similarity require all vectors to be normalized~\cite{indexbased, faiss, simplexrangesearching} to achieve good performance in practice, as they are designed under this assumption.
However, keys in transformer attention are not normalized; as we show in Appendix~\ref{app:key-norms}, their norms vary significantly.
Enforcing normalization to fit these indexes effectively alters the key representations that the model was trained with, thereby influencing attention scores and introducing additional errors at inference time.
Finally, the third key difference concerns the query-key distribution shift.
Similarity search assumes that queries and points come from the same
distribution, or at least the same nature.
In transformer attention, however, queries and keys are projections of
the same token representations through \emph{different} weight matrices,
and they live in fundamentally different regimes.
This distribution shift is already noted in prior
work~\cite{retrievalattention} and remains a challenge for methods that
directly adapt similarity search indexes to KV cache retrieval.
Halfspace range searching, by contrast, makes no such assumption: the
query $q_t$ and the key points $\mathcal{K}$ are treated as
geometrically distinct objects. 

\paragraph{Index Design Challenges.}
Classic data structures for range searching are not suitable for LLM inference due to limitations, such as high computational and update overhead.
In Appendix~\ref{sec:challenges}, we identify the three challenges that led our index design choices: (Challenge 1) high pruning power with full recall, (Challenge 2) dynamic and query-adaptive, and (Challenge 3) efficiency and integration.


\vspace{-4mm}
\section{Louver: An Inference-Efficient Index for KV Cache}

\begin{wrapfigure}{r}{0.6\linewidth}
    \centering
    \vspace{-8mm}
    \includegraphics[width=\linewidth]{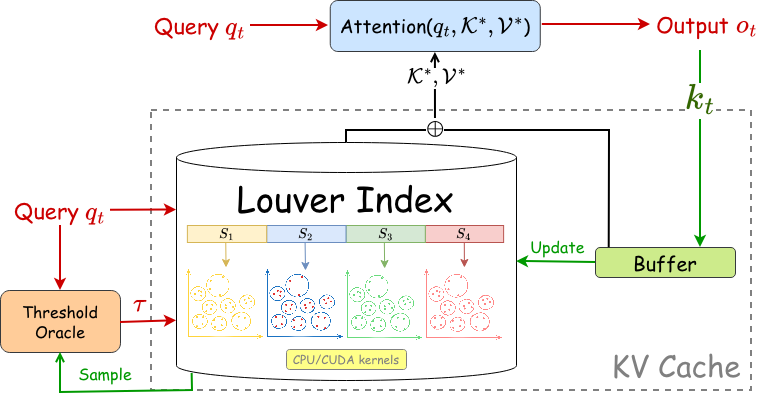}
    \vspace{-5.5mm}
    \caption{Louver index architecture.}
    \label{fig:diagram}
    \vspace{-5.5mm}
\end{wrapfigure}

In this section, we present our index, \textsc{Louver}, for KV cache retrieval. We first describe its main components, including index construction, query processing, and dynamic updates. We then analyze the index and show how it addresses the three challenges outlined above. Finally, we discuss the system-level components and implementation details. An overview of the system architecture is illustrated in Figure~\ref{fig:diagram}.

\paragraph{Overview.}
\textsc{Louver} partitions the $d$-dimensional key space into $S$ contiguous subspaces of approximately equal width and builds an independent index for each.
Within each subspace, keys are grouped into clusters of fixed size $r = O(1)$ using a balanced PCA tree, and each cluster is enclosed by a bounding ball (center + radius).
The fixed group size enables efficient hardware- and kernel-level optimizations.
At query time, given a threshold $\tau$, \textsc{Louver} tests each ball against the query halfspace in each subspace.
Clusters whose balls do not intersect the halfspace are pruned; only keys belonging to clusters that survive in all $S$ subspaces proceed to an exact dot-product check against $\tau$.
This two-phase filter guarantees zero false negatives while avoiding dot-product computation on the large majority of keys.

Keys generated during decoding are held in a small buffer $\mathcal{B}$ and attended densely at every step.
Once $|\mathcal{B}|$ reaches $B$, the buffered keys are clustered into $B/r$ new groups and appended to the index.
On GPU, this update runs on a side stream concurrently with attention; on CPU the cost amortizes to $O(1)$ per decoding step.

\vspace{-4mm}
\subsection{Index Construction}
\label{sec:index-construction}
\vspace{-3mm}

\begin{wrapfigure}{r}{0.4\linewidth}
    \centering
    \vspace{-42pt}
    \includegraphics[width=\linewidth]{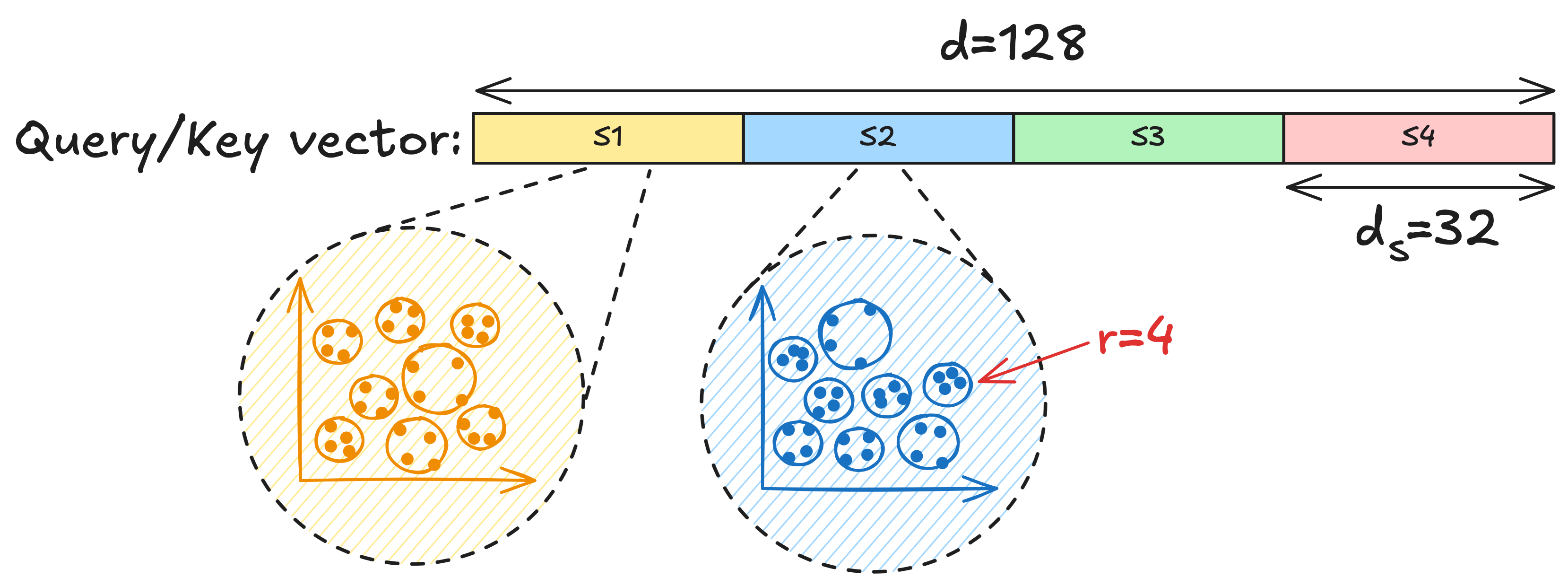}
    \vspace{-7mm}
    \caption{Subspace decomposition in \textsc{Louver}.}
    \label{fig:subspace}
    \vspace{-4mm}
\end{wrapfigure}

Given the KV cache $\mathcal{K} = \{k_1, \ldots, k_n\} \subset \mathbb{R}^d$ after prefilling, \textsc{Louver} builds a two-level index that decomposes the key space into $S$ subspaces and, within each subspace, groups keys into bounding balls of fixed size $r$.

\paragraph{Subspace decomposition.}
We partition the $d$-dimensional space into $S$ consecutive subspaces of equal size. Each subspace corresponds to a contiguous block of coordinates. For each subspace $s \in \{1,\cdots,S\}$, let $\pi_s : \mathbb{R}^d \to \mathbb{R}^{d_s}$ denote the projection onto that subspace, where $d_s = d/S$.
The index for each subspace is built independently, enabling parallel construction and sublinear query cost per subspace. This design also provides effective pruning within each subspace, as discussed in Appendix~\ref{app:ablation}.

\paragraph{Grouping via balanced PCA trees.}
Within each subspace $s$, we partition the $n$ projected keys $\{\pi_s(k_j)\}_{j=1}^n$ into $K = \lceil n/r \rceil$ groups of fixed size $r$, following the grouping definition in Appendix~\ref{app:ablation}.
We construct this partition using a \emph{balanced PCA tree}: starting from all $n$ keys, the tree recursively bisects the current point set by projecting onto the coordinate axis of maximum variance and splitting the sorted sequence at the median, distributing points equally to left and right subtrees.
The recursion terminates when a subtree contains exactly $r$ points, which form a single group.
Let $a_{s,j} \in \{1, \ldots, K\}$ denote the cluster assignment of key $k_j$ in subspace $s$, and let $\mathcal{C}_{s,i} = \{j : a_{s,j} = i\}$ be the index set of group $i$ in subspace $s$.

\paragraph{Ball enclosing.}
Each group $\mathcal{C}_{s,i}$ is enclosed\footnote{See Definition~\ref{def:1} in Appendix~\ref{app:ablation}.} by a bounding ball $B(c_{s,i}, \rho_{s,i})$.
The center is the mean of the group's projected keys:
\(
    c_{s,i} \;=\; \frac{1}{|\mathcal{C}_{s,i}|} \sum_{j \in \mathcal{C}_{s,i}} \pi_s(k_j),
\)
and the radius is the maximum distance from the center to any group member:

\vspace{-8mm}
\begin{equation}
    \hspace{15mm}\rho_{s,i} \;=\; \max_{j \in \mathcal{C}_{s,i}} \bigl\| \pi_s(k_j) - c_{s,i} \bigr\|_2.
    \label{eq:radius}
\end{equation}
\vspace{-5mm}

For a halfspace query $\mathrm{H}(q, \tau)$, the ball $B(c_{s,i}, \rho_{s,i})$ intersects $\mathrm{H}(\pi_s(q), \tau_s)$ if and only if

\vspace{-5mm}
\begin{equation}
    \langle \pi_s(q),\, c_{s,i} \rangle + \rho_{s,i}\,\|\pi_s(q)\|_2 \;\geq\; \tau_s,
    \label{eq:ball-halfspace}
\end{equation}
This calculation requires $O(d_s)$ operations per ball, regardless of the number of keys inside it.
Groups whose balls do not satisfy~\eqref{eq:ball-halfspace} are pruned entirely with no false negatives (Challenge~1), since no point inside a non-intersecting ball can satisfy the threshold.

\paragraph{Index layout and memory.}
For each subspace $s$, the index stores: the $K$ center vectors, the $K$ radii, and the $n$ assignment labels.
The additional metadata is bounded by $n/r$ descriptors per subspace — at most $1/r$ of the KV cache footprint (no more than $25\%$ for $r=4$, $S=4$).
Since the filter phase accesses only the $K = n/r$ centers, the full key array can be offloaded to slower memory tiers and fetched only for the surviving candidate set, reducing memory bandwidth pressure (see experiments in Section~\ref{sec:experiments}).
The index construction has a time complexity of $O(S \cdot n \log(n/r))$.
\\
The full procedures are provided in Algorithms~\ref{alg:build} and~\ref{alg:pca-tree} in Appendix~\ref{app:pseudocodes}.

\vspace{-4mm}
\subsection{Query Processing}
\label{sec:query-processing}
\vspace{-2mm}

At each decoding step $t$, given a query $q_t \in \mathbb{R}^d$ and a threshold $\tau$, the goal is to find $\mathcal{K}^* = \{k \in \mathcal{K} \mid \langle q_t, k \rangle \geq \tau\}$.
\textsc{Louver} supports two query algorithms that differ in how the threshold is specified and how the index is traversed.
Both share a common two-phase structure: a fast \emph{index filter} that prunes groups using cheap ball tests, followed by an \emph{exact dot-product check} on the surviving candidates to enforce the threshold precisely.
The threshold $\tau$ (or per-subspace thresholds $\tau_1, \ldots, \tau_S$) can be provided by existing thresholding methods from the literature~\cite{twilight, tactic, blasst}. Our system also includes built-in threshold modules. We design and evaluate several thresholding strategies, which are discussed in detail in Appendix~\ref{app:threshold}.

\paragraph{Query 1: Full-subspace ball filter (per-subspace thresholds).}
This variant takes $S$ per-subspace thresholds $\tau_1, \ldots, \tau_S$ as input, where $\tau_s$ is a lower bound on $\langle \pi_s(q), \pi_s(k) \rangle$ for any relevant key $k$.
Independently for each subspace $s$, a cluster $i$ is \emph{pruned} if its bounding ball does not intersect the halfspace $\mathrm{H}(\pi_s(q), \tau_s)$:

\vspace{-10mm}
\begin{equation}
    \langle \pi_s(q),\, c_{s,i} \rangle + \rho_{s,i}\,\|\pi_s(q)\|_2 \;<\; \tau_s.
    \label{eq:ball-gate-s}
\end{equation}
\vspace{-6mm}

A key $k$ is a candidate only if its cluster survives the ball test in \emph{all} $S$ subspaces.
Candidates then undergo an exact dot-product check: only keys with $\langle q, k \rangle \geq \tau$ are included in the attention sum.
Algorithm~\ref{alg:query-full} (Appendix~\ref{app:pseudocodes}) summarizes this procedure.

\paragraph{Query 2: TA filter (single global threshold).}

\begin{wrapfigure}{r}{0.5\linewidth}
    \centering
    \vspace{-10mm}
    \includegraphics[width=\linewidth]{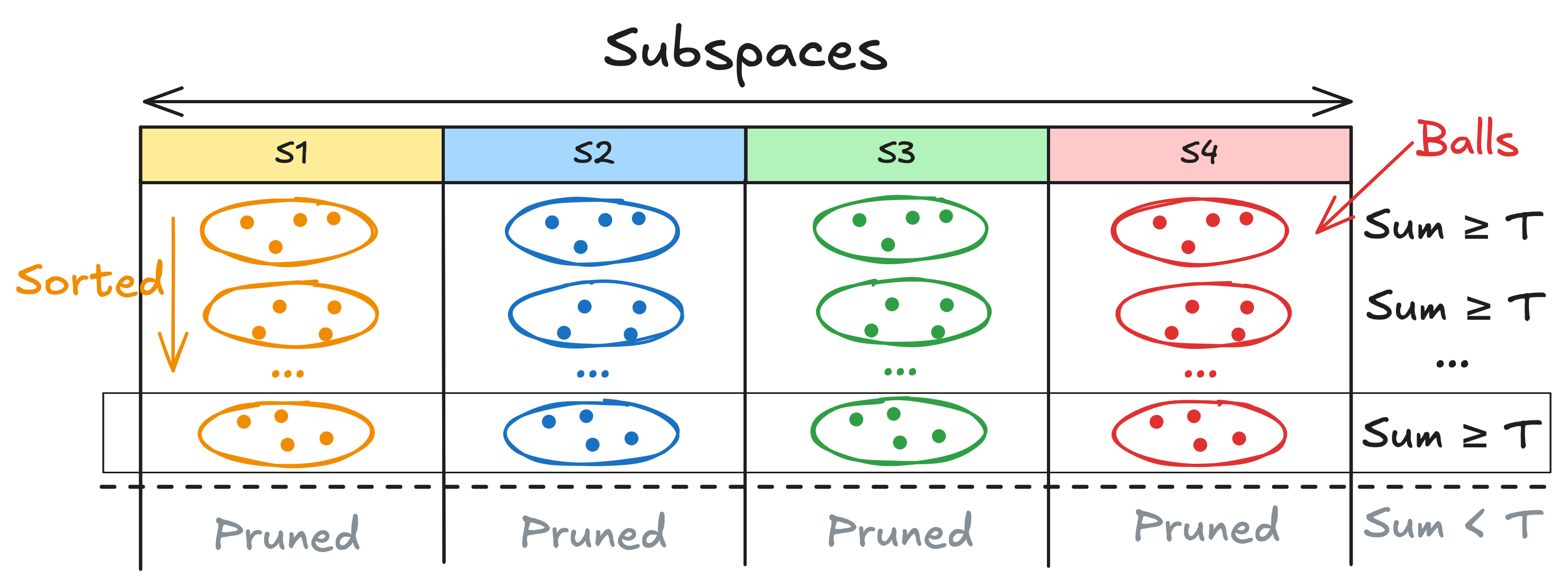}
    \vspace{-7mm}
    \caption{Visual illustration of the TA filtering procedure. Balls in each subspace are first sorted, and a cutoff is determined at the first level (row) where the aggregated score falls below the threshold.}
    \label{fig:tafilter}
    \vspace{-7mm}
\end{wrapfigure}

This variant uses a single global threshold $\tau$ and adapts the Threshold Algorithm (TA)~\cite{fagin, faginta} — originally a top-$k$ retrieval algorithm — to our threshold-based setting.

The key observation is that the full dot product decomposes across subspaces: $\langle q, k \rangle = \sum_{s=1}^S \langle \pi_s(q), \pi_s(k) \rangle$.
For any key $k$ in cluster $i$ of subspace $s$, the subspace contribution is bounded above by the ball:
\(
    \langle \pi_s(q), \pi_s(k) \rangle \;\leq\; \langle \pi_s(q), c_{s,i} \rangle + \rho_{s,i}\,\|\pi_s(q)\|_2 \;=:\; f_{s,i}.
\)
Thus $f_{s,i}$ is an upper bound on how much any key in cluster $i$ can contribute from subspace $s$.

The algorithm proceeds in rounds.
\emph{Initialization}: for each subspace $s$, compute $f_{s,i}$ for all $K$ clusters and sort them in descending order; let $\sigma_s(d)$ denote the $d$-th cluster in this order.
\emph{Each round $d$}: visit the $d$-th cluster in every subspace simultaneously and add all their keys to a live set $\mathcal{L}$.
After round $d$, the global upper bound on the full dot product of any key \emph{not yet seen in any subspace} is
\(
    U(d) \;=\; \sum_{s=1}^S f_{s,\,\sigma_s(d)}.
\)

\emph{Stopping}: scanning halts at the first depth $d^*$ where $U(d^*) < \tau$.
At this point, every key outside $\mathcal{L}$ belongs to a cluster ranked below $d^*$ in every subspace, so its full dot product cannot exceed $U(d^*) < \tau$ and is safely pruned.
Finally, keys in $\mathcal{L}$ undergo an exact dot-product check and those with $\langle q, k \rangle \geq \tau$ are included in the attention sum.
Algorithm~\ref{alg:query-ta} (Appendix~\ref{app:pseudocodes}) gives the full procedure.

\paragraph{Correctness.}
Both algorithms guarantee zero false negatives.
For Query~1: if cluster $i$ is pruned in any subspace $s$, then $\langle \pi_s(q), c_{s,i} \rangle + \rho_{s,i}\|\pi_s(q)\|_2 < \tau_s$, so no key inside the ball can achieve $\langle \pi_s(q), \pi_s(k) \rangle \geq \tau_s$, and hence cannot satisfy $\langle q, k \rangle \geq \tau$.
For Query~2: at stopping depth $d^*$, any key $k \notin \mathcal{L}$ belongs to an unseen cluster in every subspace, so $\langle q, k \rangle \leq U(d^*) < \tau$.
In both cases the final dot-product check is exact, so no false negatives are introduced.
In our experiments, the index filter prunes up to $90\%$ of keys before the exact check, meaning fewer than $10\%$ of keys reach the dot-product computation (see Section~\ref{sec:experiments}).

\vspace{-4mm}
\subsection{Dynamic Updates}
\label{sec:dynamic-updates}
\vspace{-3mm}

During decoding, a new key $k_t$ is generated at every step.
Rather than rebuilding the index from scratch, \textsc{Louver} maintains a fixed-size \emph{buffer} $\mathcal{B}$ of recently generated keys not yet incorporated into the index.
Keys in $\mathcal{B}$ are always attended densely, ensuring full recall regardless of buffer state.
\\
Once $|\mathcal{B}|$ reaches $B$, an incremental update is triggered: the same index construction procedure is applied to the $B$ buffered keys, producing $B/r$ new groups whose centers, radii, and assignments are appended directly to the main index.
The buffer is then reset.
On GPU, this update runs concurrently with the attention computation, hiding its cost; on CPU it is amortized to $O(S \cdot d_s / r)$ per step -- constant in $n$ and negligible relative to dense attention. See Appendix~\ref{app:update} for implementation details.

\paragraph{Other System Details.} 
Beyond the \textsc{Louver} index, our system (Figure~\ref{fig:diagram}) includes additional components: 
a {\em threshold selection} module that {\em automatically} identifies the relevance threshold (explained in Appendix~\ref{app:threshold}), and the {\em kernel-level implementations} optimized for GPU and CPU implementation in C++ with AVX/FP16 intrinsics (explained in Appendix~\ref{app:kernels}.
As a result, as shown in Section~\ref{sec:experiments}, \textsc{Louver} {\em outperforms even highly optimized dense attention implementations such as FlashAttention~\cite{flashattention2}}.
In Appendix~\ref{sec:louver-analysis}, we show that {\sc Louver} addresses all design challenges introduced earlier.

\begin{table}[t]
\centering
\setlength{\tabcolsep}{4pt}
\small

\caption{LongBench v1 at 10\% KV retention, Llama-3.1-8B-Instruct. F1 (\%).}
\label{tab:accuracy_longbench}
\resizebox{\textwidth}{!}{%
\begin{tabular}{llccccccc}
\toprule
\textbf{Category} & \textbf{Method} & NarrQA & Qasper & MFQ-EN & HotpotQA & 2WikiMQA & MuSiQue & \textbf{Avg} \\
\midrule
\rowcolor{gray!12}
Dense    & FlashAttention-2   & 16.4 & 46.4 & 55.0 & 54.5 & 47.1 & 30.8 & 41.7 \\
\midrule
\multirow{2}{*}{Eviction}
       & StreamingLLM       & 15.1 & 26.6 & 30.7 & 45.6 & 41.1 & 18.6 & 29.6 \\
       & H$_2$O             & 15.9 & 43.4 & 53.7 & 54.3 & 47.2 & 29.1 & 40.6 \\
\midrule
Retrieval& Quest              & 15.4 & 47.1 & 54.3 & 53.5 & 47.5 & 28.7 & 41.1 \\
\midrule
Adaptive & Twilight           & 15.7 & 46.7 & 55.0 & 54.5 & 48.2 & 29.7 & 41.6 \\
\midrule
\rowcolor{blue!18}
\textbf{Ours}   & \textbf{\textsc{Louver}}    & {\bf 16.2} & 47.1 & \textbf{55.2} & \textbf{54.5} & 47.1 & {\bf 30.9} & \textbf{41.8} \\
\bottomrule
\end{tabular}}


\begin{minipage}[t]{0.52\textwidth}
\centering
\captionof{table}{RULER at 32k tokens, 10\% KV retention, Llama-3.1-8B-Instruct. Accuracy (\%).}
\label{tab:accuracy_ruler}
\renewcommand{\arraystretch}{1.02}
\resizebox{\linewidth}{!}{%
\begin{tabular}{llcccc}
\toprule
\textbf{Category} & \textbf{Method} & NIAH-S & NIAH-M & VT & \textbf{Avg} \\
\midrule
\rowcolor{gray!12}
Dense     & FlashAttn-2   & 100.0 & 96.0 & 84.0 & 93.3 \\
\midrule
\multirow{2}{*}{Eviction}
          & StreamingLLM  &   0.0 &  9.3 &  0.0 &  3.1 \\
          & H$_2$O        & 100.0 & 96.0 & 52.0 & 82.7 \\
\midrule
Retrieval & Quest         & 100.0 & 96.0 & 60.0 & 85.3 \\
\midrule
\rowcolor{blue!18}
\textbf{Ours} & \textbf{\textsc{Louver}} & \textbf{100.0} & \textbf{98.0} & {\bf 74.0} & \textbf{90.7} \\
\bottomrule
\end{tabular}}
\renewcommand{\arraystretch}{1}
\end{minipage}%
\hfill
\begin{minipage}[t]{0.45\textwidth}
\centering
\captionof{table}{Reasoning benchmarks, DeepSeek-R1-Distill-Llama-8B. Accuracy (\%).}
\label{tab:accuracy_reasoning}
\resizebox{\linewidth}{!}{%
\begin{tabular}{llcc}
\toprule
\textbf{Category} & \textbf{Method} & \textbf{AIME} & \textbf{MATH-500} \\
\midrule
\rowcolor{gray!12}
Dense     & FlashAttn-2             & 30.0 & 58.0 \\
\midrule
\multirow{2}{*}{Eviction}
          & —            &  —   &  —   \\
          & H$_2$O                  &  —   & 55.0 \\
\midrule
Adaptive  & Twilight ($p{=}0.85$)   & 25.0 & 44.0 \\
\midrule
\rowcolor{blue!18}
\textbf{Ours} & \textbf{\textsc{Louver}} & \textbf{30.0} & \textbf{62.0} \\
\bottomrule
\end{tabular}}
\end{minipage}
\vspace{-7mm}
\end{table}

\vspace{-5mm}
\section{Experiments}\label{sec:experiments}
\vspace{-4mm}
In this section, we show the performance of our proposed index, \emph{Louver}, across multiple benchmarks and compare it under different scenarios and tasks. Additional details on experiments are provided in Appendix~\ref{app:experiments}.
We evaluate on five open-weight instruction-tuned models (full details are provided in Appendix~\ref{app:setup}): 
\textbf{Llama-3.1-8B-Instruct}~\cite{llama3}, 
\textbf{DeepSeek-R1-Distill-Llama-8B}~\cite{deepseekr1}, 
\textbf{Llama-3.2-3B-Instruct}~\cite{llama3}, 
\textbf{Qwen2.5-7B-Instruct}~\cite{qwen25}, and 
\textbf{Qwen2.5-14B-Instruct}~\cite{qwen25}. 
Our evaluation spans both long-context understanding and reasoning-intensive tasks. 
For long-context evaluation, we use LongBench~\cite{longbench} and RULER~\cite{ruler}. 
For reasoning evaluation, we use AIME 2024~\cite{aime2024} and MATH-500~\cite{math500}.
We compare \emph{Louver} against three categories of baselines. 
\emph{Fixed-budget} methods, including H$_2$O~\cite{h2o}, StreamingLLM~\cite{streamingllm}, and Quest~\cite{quest}. 
\emph{Adaptive-budget} methods, represented by Twilight~\cite{twilight}. \emph{Retrieval-based offloading} methods that maintain a retrieval index in CPU memory and perform sparse attention on the GPU.
This category includes RetrievalAttention~\cite{retrievalattention}, InfLLM~\cite{infllm}, and MagicPIG~\cite{magicpig}.
As dense-attention references, we include Torch Eager/SDPA and FlashAttention-2~\cite{flashattention2} on both GPU and CPU.
The code is publicly available in \url{https://github.com/UIC-InDeXLab/Louver}.

\vspace{-2mm}
\stitle{Sparse Attention Accuracy Across Different Tasks}
\emph{Long-context benchmarks (fixed budget):}
Tables~\ref{tab:accuracy_longbench} and~\ref{tab:accuracy_ruler} report results on LongBench and RULER using Llama-3.1-8B-Instruct under a fixed 10\% KV budget. 
\textsc{Louver} estimates a retrieval threshold from a small reservoir sample to match the same retrieval budget as the baselines. 
\textsc{Louver} achieves the best overall performance, benefiting from its zero-false-negative guarantee: all keys with attention score above $\tau$ are retrieved, preventing important context tokens from being discarded. 
In contrast, fixed-fraction methods such as Twilight can still incur false negatives under the same budget constraint.

\emph{Long-output reasoning (adaptive budget):}
Table~\ref{tab:accuracy_reasoning} reports results on AIME 2024 and MATH-500 using DeepSeek-R1-Distill-Llama-8B. 
Under continuously growing KV caches during chain-of-thought decoding, eviction and adaptive-budget baselines suffer from false negatives and accuracy degradation. 
In contrast, \textsc{Louver-TA} matches dense-attention accuracy on AIME 2024 and surpasses it on MATH-500, highlighting the importance of avoiding false negatives in long-output reasoning.

\begin{figure}[t]
\centering
\includegraphics[width=0.9\textwidth]{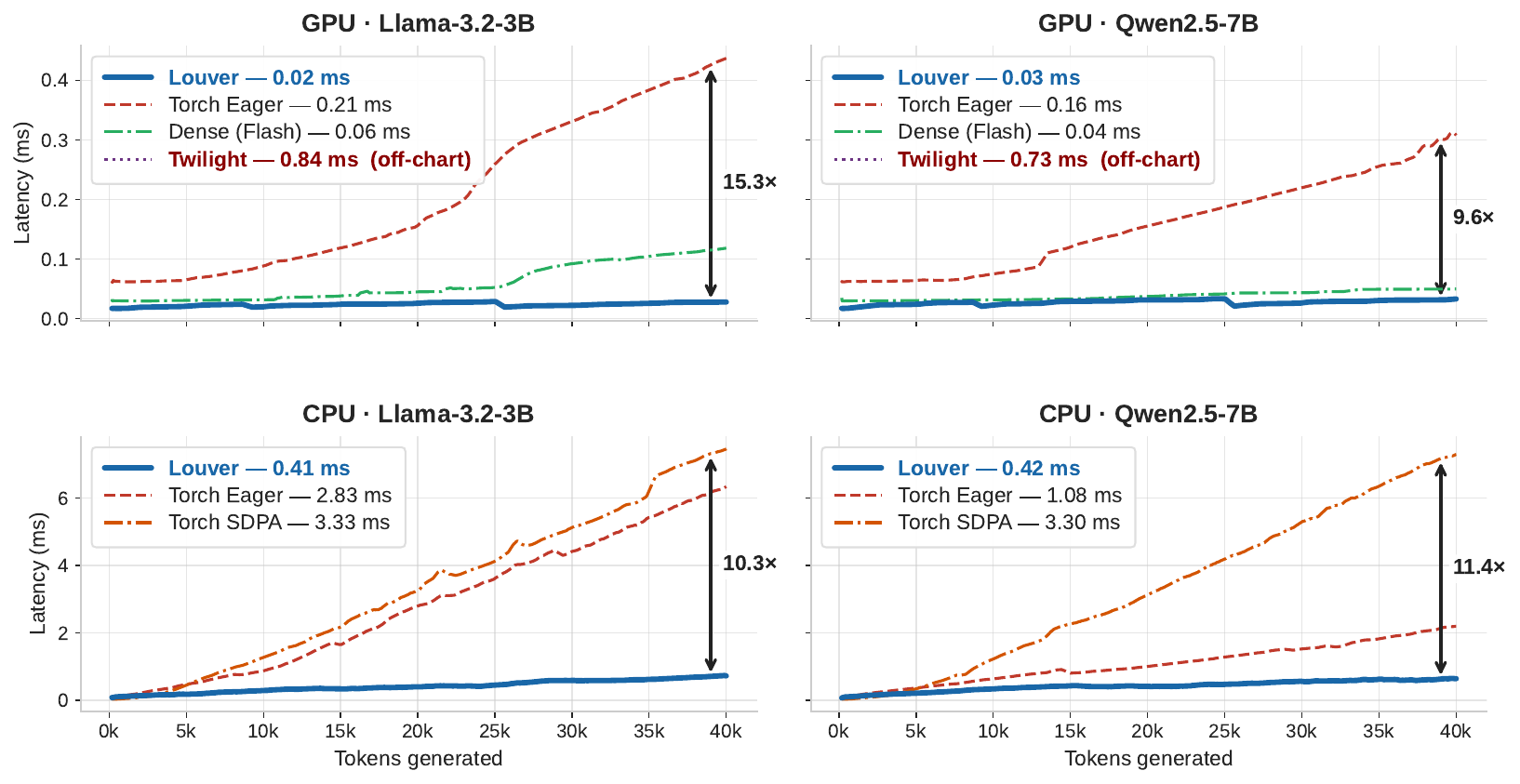}
\vspace{-2mm}
\caption{Per-step attention latency vs.\ context length. Top row: GPU; bottom row: CPU. Speedup annotations show \textsc{Louver} vs.\ the fastest dense baseline at 40k tokens.}
\label{fig:latency}
\vspace{-4mm}
\end{figure}

\vspace{-2mm}
\stitle{Long-output Decoding Latency}\label{sec:latency}
Figure~\ref{fig:latency} reports per-step attention latency during long reasoning decoding on AIME. 
Dense baselines and Twilight scale linearly with context length, since Twilight still computes full QK attention scores before pruning. 
In contrast, \textsc{Louver} prunes most keys before exact dot-product computation, resulting in sub-linear growth. 
\textsc{Louver} uses fused GPU and CPU kernels for efficient filtering and sparse attention execution. 
At 40k tokens, it achieves up to \textbf{15.3$\times$} speedup over Torch Eager on GPU and \textbf{10.3$\times$} over Torch SDPA on CPU, while maintaining dense accuracy on long-reasoning tasks.

\begin{figure}[t]
\centering
\includegraphics[width=0.9\textwidth]{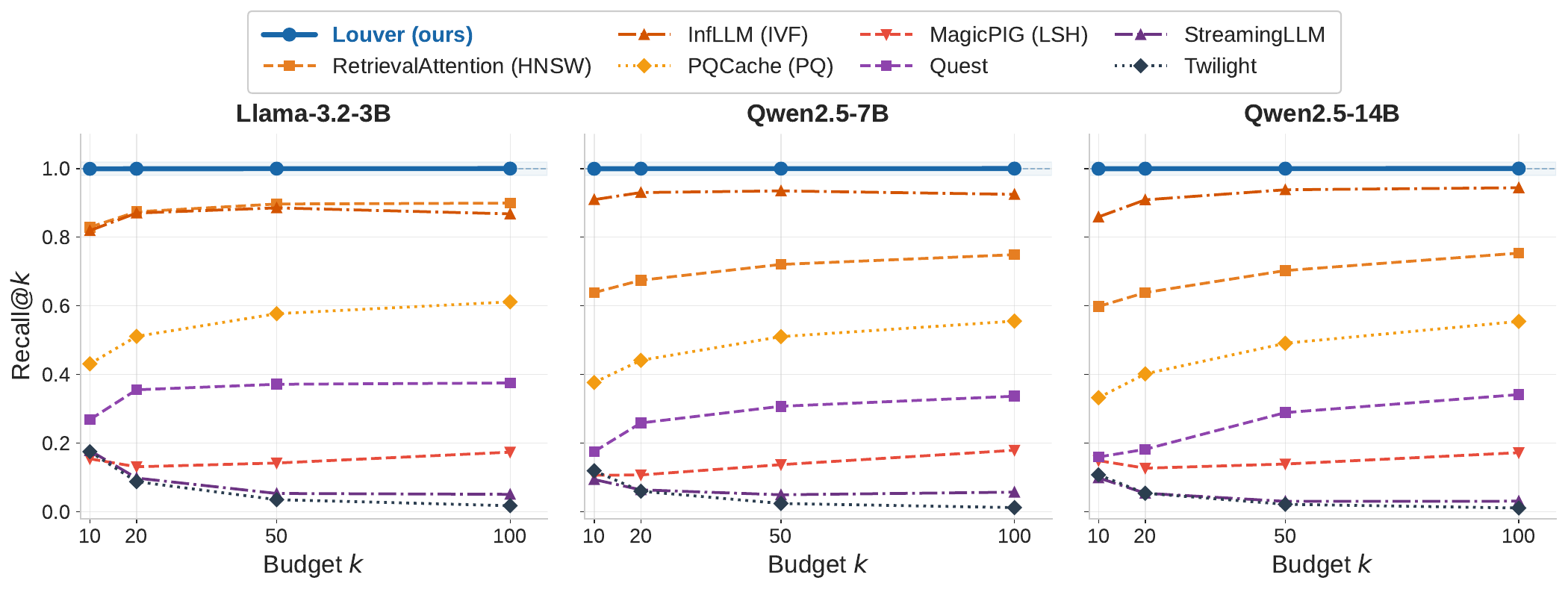}
\vspace{-2mm}
\caption{Recall@$k$ across three models. \textsc{Louver} achieves perfect recall ($\geq 99.9\%$) in all settings. Other methods fall substantially short.}
\label{fig:recall}
\vspace{-6mm}
\end{figure}

\vspace{-2mm}
\stitle{Recall Guarantee: Zero False Negatives in Practice}
We evaluate recall in two settings: ANN-based KV offloading methods and GPU-resident sparse attention methods. 
Recall is measured as the fraction of exact top-$k$ attention keys recovered during decoding, with all methods compared under the same effective token budget. 
Figure~\ref{fig:recall} shows that \textsc{Louver} consistently achieves perfect recall ($\geq 99.9\%$) across all models and values of $k$, empirically validating its zero-false-negative guarantee. 
In contrast, ANN-based retrieval methods achieve only 60--93\% recall, while sparse attention baselines perform substantially worse, often recovering below 40\% of the true top-$k$ keys.

\vspace{-2mm}
\stitle{KV Cache Offloading}\label{sec:offloading}
In the CPU-offloading setting, all methods store the full KV cache in CPU memory and retrieve only a subset of tokens to GPU at each decoding step to reduce GPU memory usage. 
In this configuration, \textsc{Louver} keeps only parent cluster centers on GPU while storing all child entries (full KV pairs) on CPU, executing its range search filtering on GPU to filter balls. 
Table~\ref{tab:offload} shows that \textsc{Louver} achieves substantially higher accuracy (38.9\% average F1) than all baselines ($\leq$26.2\%).
Its GPU-resident parent structure reduces retrieval latency to 0.07\,ms per step while the more precise retrieval prunes over 75\% of keys. 
Per-task results are provided in Appendix~\ref{app:offload_detail}.
\vspace{-4mm}

\begin{table}[h]
\centering
\small
\caption{KV offloading on LongBench avg.\ F1 (\%), Llama-3.1-8B-Instruct. Search time is index lookup per step; transfer time is CPU$\to$GPU movement of retrieved KVs.}
\label{tab:offload}
\setlength{\tabcolsep}{5pt}
\begin{tabular}{lcccc}
\toprule
\textbf{Method} & \textbf{Avg.\ F1 (\%)} & \textbf{Search (ms)} & \textbf{Transfer (ms)} & \textbf{Extra GPU (MB, \% of KV)} \\
\midrule
RetrievalAttention        & 25.1 & 53.95 & 7.05 &   0\,(0\%) \\
InfLLM                   & 26.2 &  4.89 & 7.08 & 122\,(12\%) \\
MagicPIG                 & 25.2 &  9.76 & 6.99 &  41\,(4\%) \\
\midrule
\rowcolor{blue!18}
\textbf{\textsc{Louver (offloaded)}} & \textbf{38.9} & \textbf{0.07} & \textbf{6.02} & 285\,(28\%) \\
\bottomrule
\end{tabular}
\end{table}
\vspace{-2mm}
\paragraph{Index design and threshold oracle.}
We ablate the index design and threshold-oracle variants, analyzing their impact on pruning efficiency, speedup, and accuracy--sparsity trade-offs. 
Increasing the number of subspaces improves pruning efficiency; with contiguous grouping and $S{=}16$, \textsc{Louver} scans only 16.3\% of keys while achieving a 2.42$\times$ speedup. 
Additional details are provided in Appendix~\ref{app:ablation} and Appendix~\ref{app:threshold}. Finally, a discussion on the limitations is provided in Appendix~\ref{app:limit}.

\vspace{-5mm}
\section{Conclusion}
\vspace{-4mm}
We identify false negatives in sparse attention as a major source of accuracy degradation in long-context reasoning tasks. 
By reducing token selection to halfspace range searching, we develop \textsc{Louver}, an index with a provable zero-false-negative guarantee. 
\textsc{Louver} matches or exceeds dense-attention accuracy while achieving up to 15.3$\times$ GPU and 10.3$\times$ CPU speedups at 40k context length, outperforming existing KV offloading baselines. 
We hope this work highlights the importance of avoiding false negatives in sparse attention and provides a practical path toward efficient long-context inference.

\bibliographystyle{plainnat}
\bibliography{references}

\newpage

\appendix
\section*{Appendix}
\subsection*{Table of Content}

\noindent
\noindent

\makebox[\linewidth][l]{1.\quad Additional Experiments \dotfill\ \ref{app:experiments}}\\
\makebox[\linewidth][l]{2.\quad Related Work \dotfill\ \ref{app:related}}\\
\makebox[\linewidth][l]{3.\quad Experimental Setup: False-Negativity Sensitivity \dotfill\ \ref{app:false-negativity}}\\
\makebox[\linewidth][l]{4.\quad Experimental Setup: Per-Step Score Distribution \dotfill\ \ref{app:score-distribution}}\\
\makebox[\linewidth][l]{5.\quad Key Norms Are Not Fixed \dotfill\ \ref{app:key-norms}}\\
\makebox[\linewidth][l]{6.\quad Index Design Challenges \dotfill\ \ref{sec:challenges}}
\makebox[\linewidth][l]{7.\quad Ablation on Index Design \dotfill\ \ref{app:ablation}}
\makebox[\linewidth][l]{8.\quad How Louver Addresses the Design Challenges \dotfill\ \ref{sec:louver-analysis}}
\makebox[\linewidth][l]{9.\quad Pseudo-codes \dotfill\ \ref{app:pseudocodes}}
\makebox[\linewidth][l]{10.\quad Update Process \dotfill\ \ref{app:update}}
\makebox[\linewidth][l]{11.\quad Kernel Implementation Details \dotfill\ \ref{app:kernels}}
\makebox[\linewidth][l]{12.\quad Discussion \dotfill\ \ref{app:limit}}

\section{Additional Experiments}
\label{app:experiments}

\subsection{Experimental Setup}
\label{app:setup}

\paragraph{Models.}
We use five open-weight models, all loaded in BF16 on a single GPU.
\textbf{Llama-3.1-8B-Instruct}~\cite{llama3} (128k context window) is used for accuracy on long-input benchmarks (LongBench, RULER) and offloading experiments.
\textbf{DeepSeek-R1-Distill-Llama-8B}~\cite{deepseekr1}, distilled from DeepSeek-R1 to produce extended chain-of-thought reasoning traces, is used for long-output benchmarks (AIME 2024, MATH-500).
\textbf{Llama-3.2-3B-Instruct}~\cite{llama3} and \textbf{Qwen2.5-7B-Instruct}~\cite{qwen25} are used for latency, recall, and index ablation experiments.
\textbf{Qwen2.5-14B-Instruct}~\cite{qwen25} is additionally included in the recall experiment.

\paragraph{Benchmarks.}
\emph{Long-input benchmarks} test retrieval of relevant information from a long prefill context.
\textbf{LongBench v1}~\cite{longbench} is a multi-task question-answering benchmark covering NarrativeQA, Qasper, MultiFieldQA-EN, HotpotQA, 2WikiMultiHopQA, and MuSiQue; we report average F1.
\textbf{RULER}~\cite{ruler} is a synthetic benchmark with controlled sequence lengths that isolates retrieval difficulty; we evaluate on needle-in-a-haystack single (NIAH-S), multi-needle (NIAH-M), and variable tracking (VT) at 32k tokens.
\emph{Long-output benchmarks} test accuracy under extended chain-of-thought generation, where the KV cache grows throughout decoding.
\textbf{AIME 2024}~\cite{aime2024} consists of 30 competition mathematics problems; we report pass@1 accuracy.
\textbf{MATH-500}~\cite{math500} is a curated set of 500 problems from the MATH benchmark; we report accuracy.

\paragraph{Baselines.}
\emph{Fixed-budget.}
\textbf{H$_2$O}~\cite{h2o} is eviction-based: it retains heavy-hitter tokens by cumulative attention score plus a recency window.
\textbf{StreamingLLM}~\cite{streamingllm} is eviction-based: it keeps a fixed set of attention sink tokens plus a sliding recency window.
\textbf{Quest}~\cite{quest} is retrieval-based: it scores page-level KV chunks via a $\mathrm{sign}(q)\cdot\max(k)$ proxy and retrieves the top-$k$ chunks.
All three operate at a 10\% KV budget (fraction of the full context length).

\emph{Adaptive-budget.}
\textbf{Twilight}~\cite{twilight} is retrieval-based: it computes full QK attention scores at each step, softmax-normalizes them, and retains the smallest set of tokens whose cumulative probability mass reaches $p = 0.85$.
We use the same threshold $p = 0.85$ for \textsc{Louver-TA}, making the budget directly comparable.

\emph{Retrieval-based offloading.}
All three methods offload the full KV cache to CPU and retrieve a budget of tokens per step via an ANN index.
\textbf{RetrievalAttention}~\cite{retrievalattention} builds an HNSW index on CPU and fetches approximate nearest-neighbor KV pairs on demand.
\textbf{InfLLM}~\cite{infllm} uses IVF clustering on CPU and retrieves the top-scoring clusters per step.
\textbf{MagicPIG}~\cite{magicpig} uses LSH on CPU to identify relevant KV pairs via hash bucket lookup.
All three retrieve 15\% of stored tokens per step.

\paragraph{Dense attention.}
\textbf{Torch Eager} uses the standard unfused $O(N)$ implementation.
\textbf{FlashAttention-2}~\cite{flashattention2} is the IO-aware fused kernel and our primary dense accuracy upper bound.
\textbf{Torch SDPA} dispatches to the most efficient available backend (FlashAttention-2 or memory-efficient attention) and is used as the CPU dense reference in latency experiments.

\paragraph{System configuration.}
\emph{Accuracy experiments} (LongBench, RULER, AIME 2024, MATH-500, offloading) are run on the Delta cluster (NCSA) on a single-GPU node equipped with one \textbf{NVIDIA A100 40\,GB SXM4} (Ampere, 40\,GB HBM2e, 1.6\,TB/s memory bandwidth), 64 CPU cores (AMD EPYC 7763), and 256\,GB of RAM.

\emph{Latency and recall experiments} are run on a local workstation with an \textbf{NVIDIA RTX\,5090} (32\,GB), an \textbf{AMD Ryzen Threadripper 7970X} (64 threads), and 256\,GB DDR5.
CPU latency experiments use the Threadripper with the GPU idle.
All models are loaded in BF16; CUDA 12.4, PyTorch 2.5, and Triton 3.1 are used throughout.

\subsection{KV Cache Offloading: Per-Task Breakdown}
\label{app:offload_detail}

Table~\ref{tab:offload_pertask} reports per-task LongBench F1 scores for all offloading methods at a 15\% token budget on Llama-3.1-8B-Instruct.
\textsc{Louver} leads on every task, with the largest gains on multi-hop reasoning tasks (MuSiQue, Qasper) that require precise retrieval of specific evidence tokens — exactly the setting where its zero-false-negative guarantee matters most.

\begin{table}[h]
\centering
\small
\caption{Per-task LongBench F1 (\%) for KV offloading methods at 15\% budget, Llama-3.1-8B-Instruct.}
\label{tab:offload_pertask}
\resizebox{\textwidth}{!}{%
\begin{tabular}{llcccccc}
\toprule
\textbf{Category} & \textbf{Method} & HotpotQA & 2WikiMQA & MuSiQue & Qasper & NarrativeQA & \textbf{Avg} \\
\midrule
\multirow{3}{*}{Offloading}
  & RetrievalAttention (HNSW) & 41.0 & 27.4 & 21.3 & 17.9 & 18.0 & 25.1 \\
  & InfLLM (IVF)              & 51.0 & 19.4 & 20.0 & 19.9 & 20.5 & 26.2 \\
  & MagicPIG (LSH)            & 42.0 & 24.2 & 20.0 & 17.5 & 22.3 & 25.2 \\
\midrule
\rowcolor{blue!18}
\textbf{Ours} & \textbf{\textsc{Louver}} & \textbf{59.0} & \textbf{32.5} & \textbf{35.0} & \textbf{36.3} & \textbf{31.9} & \textbf{38.9} \\
\bottomrule
\end{tabular}}
\end{table}

\subsection{Threshold Oracle Comparison}
\label{app:threshold}

\textsc{Louver} is agnostic to how the threshold $\tau$ is obtained and accepts any oracle as input, including methods from prior work~\cite{twilight, tactic, blasst}.
We additionally provide three built-in sampling-based oracles that maintain a reservoir of past keys during decoding and estimate $\tau$ from dot products against the current query; the reservoir is updated in $O(1)$ per step via reservoir sampling.
\begin{itemize}[leftmargin=*, nosep]
  \item \textbf{Sample-Max}: $\tau = $ maximum dot product of the query against the reservoir. Conservative; tends to over-include.
  \item \textbf{Sample-MeanMax}: $\tau = $ average of sample-max and sample-mean scores. Less conservative, tighter candidate sets.
  \item \textbf{Sample-Gap}: cut at the largest gap between consecutive sorted reservoir scores. Adapts to natural score distribution discontinuities.
\end{itemize}

For the ablation below we additionally evaluate \textbf{sample-topk($m$)} ($\tau = m$-th largest reservoir score, $m\in\{2,5,10\}$) and \textbf{budget($\alpha$)} ($\tau = (1{-}\alpha)$-quantile of reservoir, targeting retrieval fraction $\alpha$), giving nine variants in total.
All are estimated from a reservoir of size 256.

\paragraph{Attention concentration.}
Attention distributions during long-output decoding are highly concentrated: on average, only \textbf{3.9\%} of keys are needed to cover 50\% of the total softmax mass, with low variance (std 0.87\%) across 2000 decode steps on DeepSeek-R1-Distill-Qwen-14B (see also Figure~\ref{fig:decoding} in the main paper).
This concentration justifies threshold-based retrieval: the important tokens form a stable, small set that a reservoir-estimated $\tau$ can reliably identify.

\begin{figure}[h]
\centering
\includegraphics[width=0.6\textwidth]{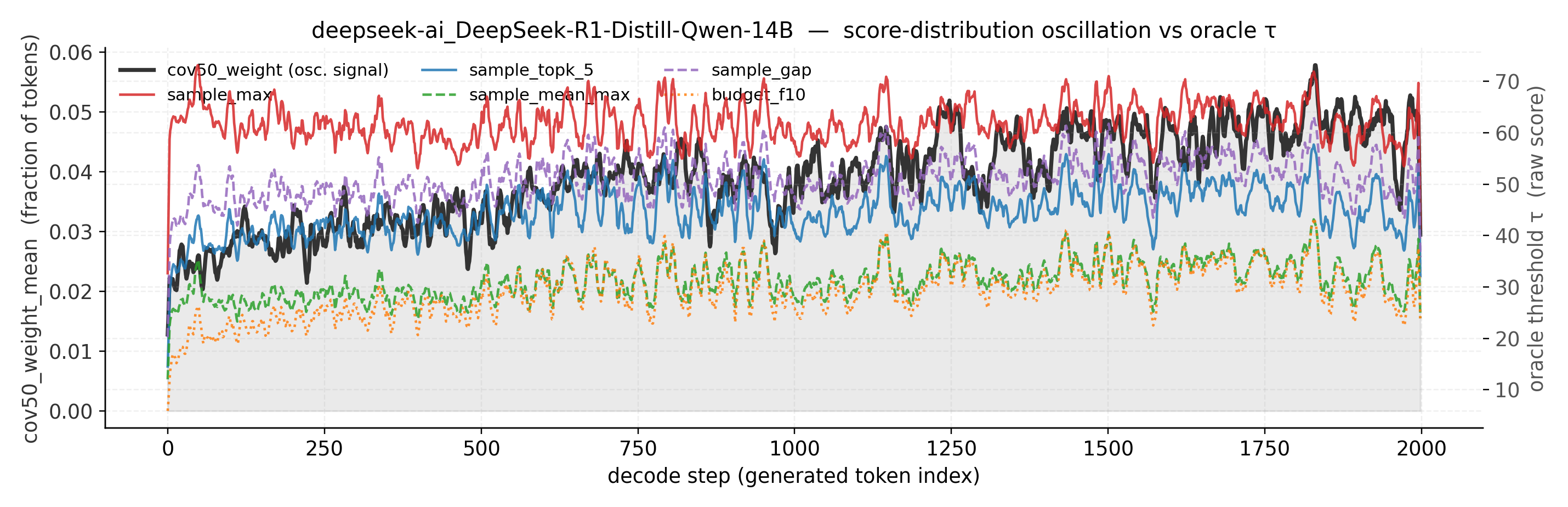}
\caption{Threshold $\tau$ trajectories for each oracle variant over 2000 decode steps (DeepSeek-R1-Distill-Qwen-14B). Sample-based oracles (sample-max, sample-gap) are most stable; budget oracles show higher variance as the score distribution shifts.}
\label{fig:oracle_oscillation}
\end{figure}

\paragraph{Oracle stability.}
Table~\ref{tab:threshold_oracle} reports the mean and standard deviation of $\tau$ across 2000 decode steps (context length 772--2770 tokens, top-25\% layers, DeepSeek-R1-Distill-Qwen-14B).
All oracles produce stable thresholds (CoV $\leq 25\%$); budget-based oracles have higher variance because they track the quantile of a shifting distribution, while sample-gap and sample-max are anchored to the reservoir's extremes and are more stable.
Figure~\ref{fig:oracle_oscillation} shows $\tau$ trajectories over decode steps for each oracle.

\begin{table}[h]
\centering
\small
\caption{Threshold oracle comparison: mean and std of $\tau$ over 2000 decode steps. DeepSeek-R1-Distill-Qwen-14B, reservoir size 256, top-25\% layers averaged.}
\label{tab:threshold_oracle}
\setlength{\tabcolsep}{6pt}
\begin{tabular}{lrrrr}
\toprule
\textbf{Oracle} & \textbf{$\tau$ mean} & \textbf{$\tau$ std} & \textbf{CoV} & \textbf{Description} \\
\midrule
sample-max          & 62.0 & 5.3 &  8.5\% & max of reservoir \\
sample-topk(2)      & 52.8 & 5.6 & 10.6\% & 2nd largest in reservoir \\
sample-gap          & 51.1 & 5.5 & 10.8\% & largest gap in reservoir \\
sample-topk(5)      & 44.8 & 5.7 & 12.7\% & 5th largest in reservoir \\
sample-topk(10)     & 38.6 & 5.7 & 14.8\% & 10th largest in reservoir \\
sample-mean-max     & 31.3 & 4.6 & 14.6\% & mean of per-head maxima \\
budget(5\%)         & 36.8 & 5.7 & 15.6\% & 95th-pct of reservoir \\
budget(10\%)        & 29.2 & 5.8 & 19.8\% & 90th-pct of reservoir \\
budget(15\%)        & 24.2 & 5.9 & 24.4\% & 85th-pct of reservoir \\
\bottomrule
\end{tabular}
\end{table}

\section{Related Work}\label{app:related}
\paragraph{Inference-time Sparse Attention and KV Cache Management.}
Sparse attention for long-context inference can be broadly divided into \emph{eviction-based} and \emph{retrieval-based} approaches.
Eviction-based methods permanently remove tokens from the KV cache to reduce memory or bandwidth cost.
Earlier approaches rely on fixed sparsity patterns such as sliding windows or attention sinks~\cite{sparsetransformer, longformer, streamingllm}, while more recent work dynamically selects tokens to discard using runtime statistics or importance estimates, including SnapKV~\cite{snapkv}, H$_2$O~\cite{h2o}, FastGen~\cite{fastgen}, Scissorhands~\cite{scissorhands}, and Mustafar~\cite{mustafar}.
Some methods instead focus on reducing attention cost during prefilling over long inputs, including MInference~\cite{minference}, FlexPrefill~\cite{flexprefill}, and XAttention~\cite{xattention}.
A common limitation of eviction-based approaches is irreversibility: once tokens are removed or compressed, they cannot be recovered later if they become relevant.

Retrieval-based methods preserve the full KV cache logically and select relevant entries dynamically at inference time.
Some systems offload the KV cache to lower tiers of the memory hierarchy and retrieve entries on demand, including RetrievalAttention~\cite{retrievalattention}, MagicPIG~\cite{magicpig}, PQCache~\cite{pqcache}, ArkVale~\cite{arkvale}, InfLLM~\cite{infllm}, FreeKV~\cite{freekv}, and ShadowKV~\cite{shadowkv}.
Others keep all entries on-device but reduce memory bandwidth by selecting only a subset of important keys during attention computation, including SparQ~\cite{sparq}, Quest~\cite{quest}, HashAttention~\cite{hashattention}, Loki~\cite{loki}, SpargeAttention~\cite{spargeattention}, and Squeezed Attention~\cite{squeezedattention}.
Our work falls in this category: we formulate sparse attention as a range-search problem over key vectors and introduce Louver, a dynamically updated index supporting exact thresholded retrieval with \emph{zero false negatives} during decoding. It also generalizes to both offloading and non-offloading settings.

\paragraph{KV Cache Indexing.}
Retrieval-based sparse attention methods differ primarily in how they index the KV cache.
Several cast the selection problem as approximate nearest neighbor search (ANN) or maximum inner product search (MIPS).
MagicPIG~\cite{magicpig} uses locality-sensitive hashing (LSH), PQCache~\cite{pqcache} uses product quantization (PQ), and RetrievalAttention~\cite{retrievalattention} constructs an attention-aware vector retrieval index tailored to the distribution mismatch between queries and cached keys.
HashAttention~\cite{hashattention} instead uses learned binary signatures in Hamming space for efficient token selection.
Squeezed Attention~\cite{squeezedattention} builds a hierarchical centroid-based index for fixed prompt contexts constructed during prefilling.
In contrast, our approach performs \emph{range searching} over key vectors without requiring normalization or dealing with query-key distributional mismatch~\cite{retrievalattention}, and supports incremental updates during decoding.

\paragraph{Adaptive Sparsity Budgets.}
Most sparse attention methods select a fixed number of tokens per query.
Recent work explores adaptive sparsity budgets based on attention mass or coverage criteria.
Twilight~\cite{twilight} selects tokens until cumulative attention mass exceeds a threshold,
Tactic~\cite{tactic} uses coverage-based stopping criteria,
and SampleAttention~\cite{sampleattention} adaptively identifies a minimal structured subset of key-value pairs.
Per-layer and per-head budget allocation are explored by PyramidInfer~\cite{pyramidinfer}, PyramidKV~\cite{pyramidkv}, and Ada-KV~\cite{adakv}.
BLASST~\cite{blasst} instead prunes low-importance attention blocks inside tiled hardware-optimized attention kernels using calibrated thresholds.
While these works address the same challenge for a fixed-size formulation, they don't guarantee the \emph{zero false negativity}. Our approach differs in that it naturally supports dynamic, per-query thresholds that vary across heads, layers, and decoding steps, while retrieving \emph{all keys} satisfying the threshold. This is a result of designing an index that handles \emph{range searching} queries.

\paragraph{Complementary Attention Acceleration Techniques.}
Hardware-aware exact-attention kernels such as FlashAttention~\cite{flashattention2} and serving systems such as vLLM~\cite{vllm} improve dense attention efficiency and are complementary to sparse retrieval methods.
Similarly, linear-complexity attention architectures and KV cache quantization techniques~\cite{surveyefficientattention, surveykvmanagement} address different aspects of long-context inference and can be combined with our approach.

\section{Experimental Setup: False-Negative Sensitivity (Observation~1)}
\label{app:false-negativity}

\paragraph{Data and prompt.}
We construct a synthetic summation task that gives us full control over which
context tokens are \emph{relevant} (i.e., must be attended to in order to
produce the correct answer). Each instance is a list of $N$ integers drawn
uniformly from $\{1,2,3\}$, formatted as the prompt
\textit{``Consider the list of numbers: $x_1, x_2, \ldots, x_N$. What is the
sum of the numbers? Answer with just one integer and nothing else. The answer
is''}. Greedy decoding produces the answer token, which is compared against
$\sum_i x_i$. We tokenise the prompt with the model's native tokenizer and
record, for each number $x_i$, the set of token positions that span its
surface form; these positions form the \emph{relevant} set $R$. Filler tokens
matching the closed-class set $\{$``the'', ``of'', ``a'', ``an'', ``and'',
``with'', ``to'', ``in'', ``on'', ``just'', ``one'', ``nothing'', ``else'',
``this'', ``,'', ``.'', ``:''$\}$ form the \emph{irrelevant} set $\bar R$.

\paragraph{Models and sparse-attention simulation.}
We evaluate Llama-3.2-1B/3B-Instruct, Llama-3.1-8B-Instruct, and
Qwen2.5-3B/7B-Instruct in \texttt{fp16} with eager attention. Sparse attention
is simulated by masking selected \emph{key} positions: we run a single
\texttt{forward} pass with a 4-D additive attention mask of shape
$(1,1,S,S)$ that combines the standard causal mask with $-\infty$ entries on
the columns of dropped positions, so that no query attends to those keys. This
is mathematically equivalent to evicting those KV-cache entries while keeping
all model weights and remaining keys/values intact. For Figure~\ref{fig:drop-error}
(Experiment~1) we drop $M{=}1$ token at a time, comparing one randomly chosen
token from $R$ with one randomly chosen token from $\bar R$, with $N{=}8$ and
$30$ random lists per model. For Figure~\ref{fig:error-spike} (Experiment~3)
we sweep $N\in\{2,3,4,5,6,8,10,12,16\}$ and a fixed-$K$ budget
$K\in\{2,3,4,6\}$ that retains exactly $K$ of the $N$ number tokens (chosen
uniformly at random) plus structural context (BOS, the question span, and all
non-list tokens), and compare against the dense baseline and an
oracle ``keep-all-relevant'' policy that retains every token in $R$. Each
configuration is averaged over $20$ random lists with seed $2$.

\paragraph{Metrics.}
We use two complementary metrics that together capture both decoding behaviour
and full distributional shift. (i) $P(\text{answer changed vs.\ dense})$ is
the fraction of trials in which greedy decoding under the dropped mask
produces a different output string than greedy decoding under dense attention,
when both are run for $4$ tokens; this is the metric reported in
Figure~\ref{fig:drop-error}. (ii) $\mathrm{KL}(p_{\text{drop}}\,\|\,p_{\text{dense}})$
is the Kullback--Leibler divergence between the next-token distributions
obtained from the dense and dropped forward passes after softmax over the full
vocabulary; this is the metric reported in Figure~\ref{fig:error-spike}. Both
are averaged across trials per $(N, \text{method})$ cell. Standard errors on
the bar plot use the binomial estimator
$\sqrt{p(1-p)/n}$.

\section{Experimental Setup: Per-Step Score Distribution (Observation~2)}
\label{app:score-distribution}

\paragraph{Model, prompt, and generation.}
We use \texttt{DeepSeek-R1-Distill-Qwen-14B} loaded in \texttt{fp16} with
eager attention, decoded greedily for $2{,}000$ new tokens on a single
multi-part reasoning prompt. The prompt asks the model to solve a six-part
combinatorial scheduling problem (assigning $7$ cities to $5$ days under
seven explicit constraints, costing the resulting schedules, deriving a
counting recurrence, identifying equivalence classes, writing pseudocode for
a backtracking solver, and recomputing counts under perturbed constraints).
Two stylistic instructions are pinned in the system prompt to elicit
representative reasoning behaviour: (A) the model must produce an explicit
numbered plan and at least four ``pivot'' blocks beginning with
\textit{``Wait,''} / \textit{``Actually,''} / \textit{``Hmm, let me reconsider~--''},
and (B) every numeric quantity must be re-derived inline as
\texttt{LHS = a OP b = result}. These instructions induce both wide-tail
synthesis steps (pivots, plan back-references, recaps) and narrow-tail local
steps (intermediate arithmetic, template scaffolding) within the same trace,
which is what the cov50 metric is designed to discriminate.

\paragraph{Q/K capture and metric.}
During decoding we hook every attention layer and persist, for each generated
position $t$, the post-RoPE query vector $q_{t,h}\in\mathbb{R}^{d}$ for every
query head $h$, alongside the full prefix of post-RoPE keys
$k_{1:T_t,h'}$ for every key/value head $h'$, where $T_t$ is the prompt
length plus $t$. The snapshot is replayed offline. For each layer we
expand the GQA mapping $h \mapsto h' = h \bmod (\text{num\_kv\_heads})$ and
compute attention scores in \texttt{fp32} as $s_{t,h,j} = q_{t,h}^{\top}
k_{j,h'} / \sqrt{d}$ for $j \le T_t$, then weights $w_{t,h,j} =
\mathrm{softmax}_j(s_{t,h,j})$. The per-head, per-step \emph{cov50} statistic
is the smallest $k$ such that the top-$k$ cumulative mass of
$(w_{t,h,j})_{j}$ reaches $0.5$, normalised by $T_t$:
$$
\mathrm{cov50}_{\text{weight}}(t,h) \;=\; \frac{1}{T_t}\,\min\Bigl\{k \;:\;
\sum_{j \in \mathrm{top}_k(w_{t,h,\cdot})} w_{t,h,j} \ge 0.5\Bigr\}.
$$
A score-space variant $\mathrm{cov50}_{\text{score}}$ is computed identically
after shifting the raw scores by their per-step minimum and renormalising to
a distribution; we report the weight-space variant in
Figure~\ref{fig:decoding}. Cov50 is bounded in $(0,1]$ and is comparable
across decoding positions of different lengths $T_t$: small values indicate
that attention concentrates on a tiny fraction of the prefix, while large
values indicate diffuse attention.

\paragraph{Aggregation and figure.}
Computation runs on the GPU, layer by layer, with chunking over the query
dimension to bound memory. For each step $t$ we aggregate
$\mathrm{cov50}_{\text{weight}}(t,h)$ across all heads of all layers by taking
the mean, yielding the time-series plotted in Figure~\ref{fig:decoding}. The
prefill region ($t < $ prompt length) is shown faded for reference; only the
decode region is annotated. To ground the visual signal, we automatically
extract decode-step windows corresponding to the four highest and four lowest
cov50 values (separated by at least $120$ tokens to avoid clustering) and
decode the $\pm 20$-token neighbourhood of each peak.

\paragraph{Decoded windows.}
We list the eight extracted windows below; the bracketed token \texttt{[[\,]]}
marks the target step $t$, and the value next to each label is the mean
cov50 at that step.

\textit{Wide-tail (transition / synthesis) windows:}
\begin{itemize}\itemsep1pt
  \item \textsc{w1} \,($t{=}1365$, $\mathrm{cov50}{=}0.058$).
    \texttt{``\,that would make Mon have two cities: G and A. Is that
    allowed? Constraint (f) [[\,says\,]] at most 2 per day, so yes. Wait,
    but in this subcase, G is\,''}
  \item \textsc{w2} \,($t{=}1460$, $\mathrm{cov50}{=}0.060$).
    \texttt{``\,can't be on Tue because that would make three cities on Tue
    (C, F, B). [[\,Wait\,]], no, each day can have up to 2 cities. So if B
    is on Tue,\,''}
  \item \textsc{w3} \,($t{=}1648$, $\mathrm{cov50}{=}0.054$).
    \texttt{``\,we have 7 cities: A, B, C, D, E, F, G. [[\,All\,]] are
    assigned except Fri. Wait, no, Fri is empty. So we need to assign the
    remaining\,''}
  \item \textsc{w4} \,($t{=}1767$, $\mathrm{cov50}{=}0.056$).
    \texttt{``\,because we have to assign all cities, but Fri is empty. So
    this is invalid. Wait, [[\,no\,]], Fri can have zero cities, but the
    problem says each city has exactly one delivery, so all\,''}
\end{itemize}

\textit{Narrow-tail (local / template) windows:}
\begin{itemize}\itemsep1pt
  \item \textsc{n1} \,($t{=}1050$, $\mathrm{cov50}{=}0.019$).
    \texttt{``\,Fri is available. So, four possibilities for G and F. Let me
    consider each subcase [[\,.\,]] \textbf{Subcase 1a: G on Mon, F on
    Tue}\,''}
  \item \textsc{n2} \,($t{=}1265$, $\mathrm{cov50}{=}0.029$).
    \texttt{``\,A can't be on Thu. So, possible A assignments: Mon, Tue,
    Wed. Let [[\,'s\,]] try A on Mon. Then B can be on Tue, Wed, Thu.\,''}
  \item \textsc{n3} \,($t{=}1557$, $\mathrm{cov50}{=}0.022$).
    \texttt{``\,has D and B. That's allowed. So, two possibilities for B.
    Let's explore both [[\,.\,]] \textbf{Subsubcase 1a1: A on Mon, B on
    Wed}\,''}
  \item \textsc{n4} \,($t{=}1822$, $\mathrm{cov50}{=}0.029$).
    \texttt{``\,cities except Fri, which is empty. That's a problem. So this
    subsubcase is invalid [[\,.\,]] Wait, no, let me recount: Mon: G, A
    (2 cities), Tue: C\,''}
\end{itemize}

\paragraph{Discussion.}
The two clusters expose qualitatively distinct attention regimes within the
\emph{same} decoding trace. Wide-tail steps consistently fire at points where
the model must integrate information from far-apart spans of the prefix:
checking a freshly-emitted partial schedule against constraint (f) declared
hundreds of tokens earlier (\textsc{w1}), pivoting away from a wrong
intermediate conclusion by re-reading the per-day capacity rule (\textsc{w2}),
recapping the seven-city universe before reasoning about the remaining slots
(\textsc{w3}), or revisiting whether ``Fri empty'' is actually permitted
(\textsc{w4}). All four contain the model's characteristic
\textit{``Wait,~no\dots''} pivot phrase and force attention to spread over
many earlier tokens to reconcile the local state with global constraints.
Narrow-tail steps, by contrast, sit at template boundaries where the next
token is essentially determined by the most recent few words: the period
that closes a clause and opens a new \textbf{Subcase} block (\textsc{n1},
\textsc{n3}), the contraction \textit{``Let's''} that introduces a fresh
local trial (\textsc{n2}), or the period that closes
\textit{``\dots is invalid''} before backtracking (\textsc{n4}). Cov50 values
differ by roughly a factor of three between the two regimes
($\sim0.02$ vs.\ $\sim0.06$), and the alternation between them happens on a
scale of tens to a few hundred decode steps. Any sparse-attention policy
with a fixed token budget either over-provisions on narrow steps (wasting
compute) or under-provisions on wide steps (incurring false negatives on the
exact tokens the model is trying to reconcile), which is the empirical basis
for the dynamic, recall-oriented formulation argued for in the main text.

\section{Key Norms Are Not Fixed}
\label{app:key-norms}

\begin{figure}[h]
\centering

\begin{minipage}{0.48\linewidth}
  \centering
  \includegraphics[width=\linewidth]{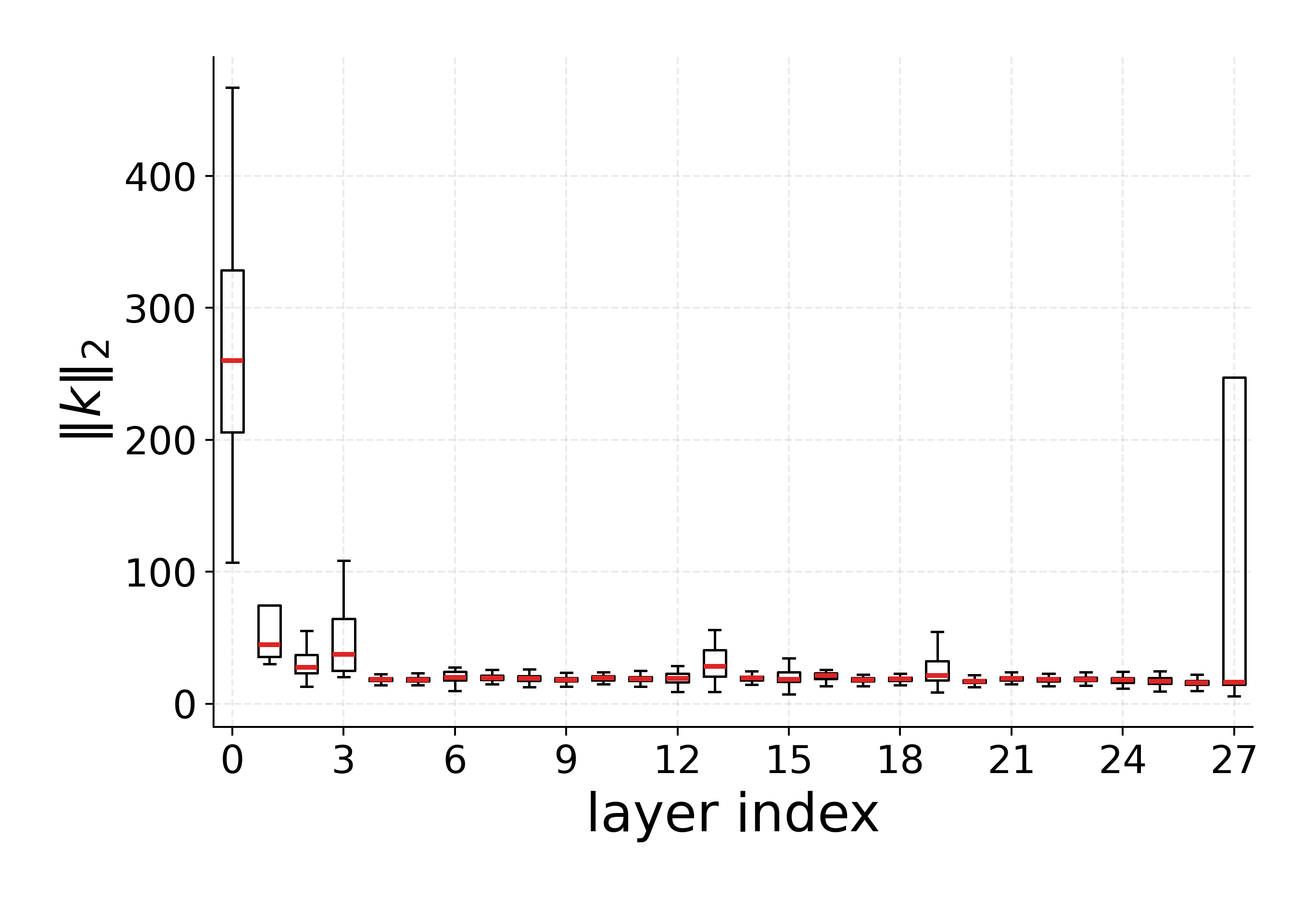}
  \caption{Per-layer distribution of key-vector L2 norms across all KV-heads
  and token positions in a $128$-token Qwen2.5-7B trace. The within-layer
  spread is wide and the median grows substantially with depth, so a single
  global norm assumption is not justified.}
  \label{fig:keynorm-box}
\end{minipage}
\hfill
\begin{minipage}{0.48\linewidth}
  \centering
  \includegraphics[width=\linewidth]{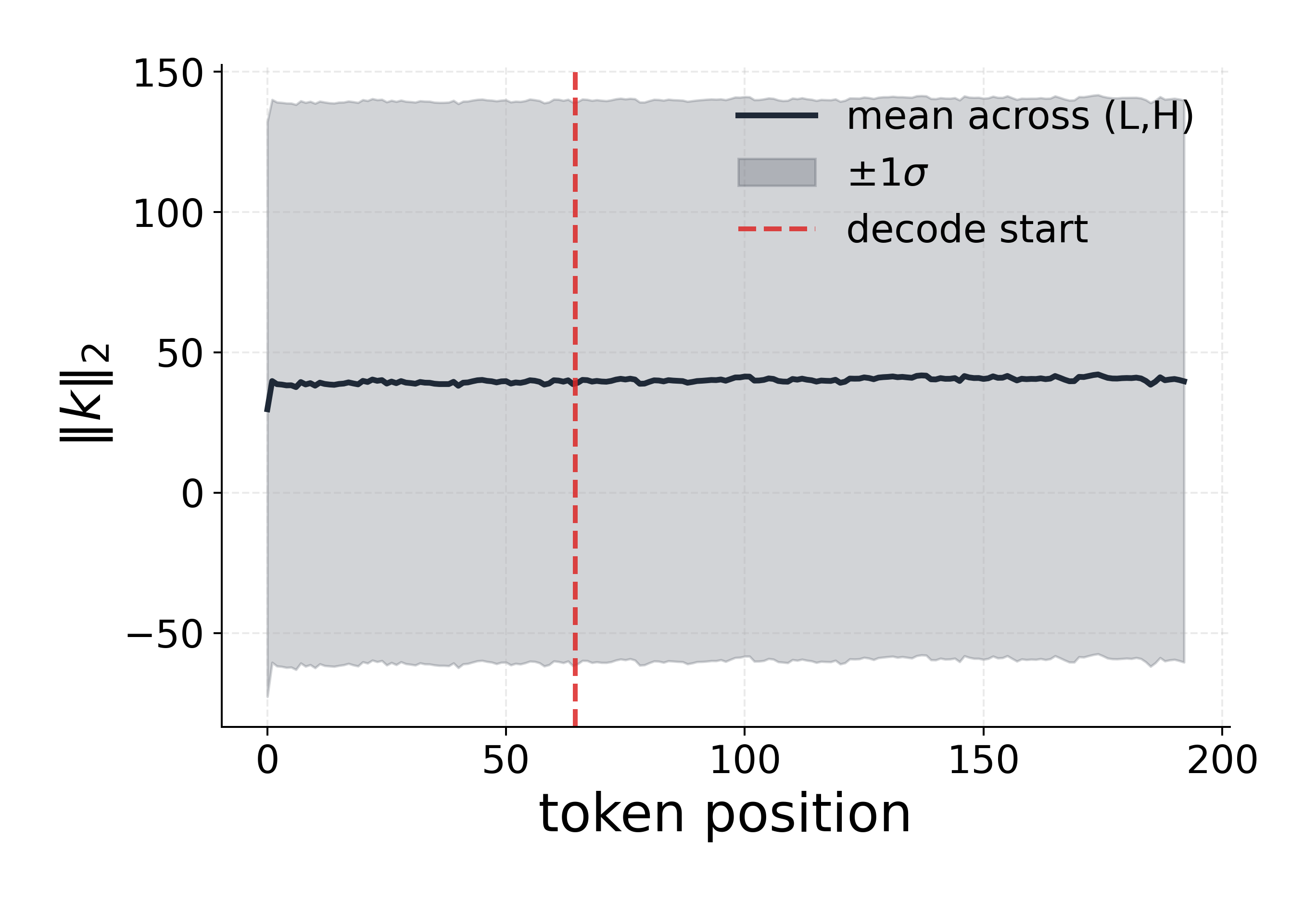}
  \caption{Mean key-norm at each token position (averaged across all
  $L\times H$ heads) with a $\pm 1\sigma$ envelope. The dashed line marks
  the prompt$\rightarrow$decode boundary. Norms peak at the first few
  positions (BOS / system header) and exhibit position-dependent drift
  rather than a constant value.}
  \label{fig:keynorm-pos}
\end{minipage}

\end{figure}

\paragraph{Setup.}
We run \texttt{Qwen2.5-7B-Instruct} (\texttt{fp16}, eager attention) on a
short technical prompt and decode $128$ greedy tokens, hooking every
attention layer to record the post-RoPE key vector $k_{\ell,h,t}$ for each
layer $\ell$, KV-head $h$, and token position $t$. We then compute the L2
norm $\|k_{\ell,h,t}\|_2$ and study how it varies across the three axes.
Aggregated over all $(L,H,T)=(28,4,193)$ entries the norm spans
$[0.37,\,923]$ with mean $40.0$ and standard deviation $100.1$; the
$\max/\min$ ratio is $\approx 2.5\times 10^{3}$, i.e.\ keys are far from
unit-normalised. Marginalising one axis at a time, the per-layer mean still
varies by $17\times$, the per-head mean by $2.5\times$, and the per-position
mean by $1.4\times$, so the variation is not noise but structured along each
axis. Figure~\ref{fig:keynorm-box} shows the within-layer distribution and
makes the depth dependence explicit; Figure~\ref{fig:keynorm-pos} shows the
mean and $\pm 1\sigma$ envelope at each position, revealing systematic
spikes at the BOS / system-header tokens and around the
prompt$\rightarrow$decode boundary.

\paragraph{Implication for sparse attention.}
Attention scores $s_{t,j} = q_t^\top k_j / \sqrt{d}$ scale linearly with
$\|k_j\|$, so the norm is a real component of relevance: a key with large
$\|k\|$ contributes more mass under any query whose direction is not
orthogonal to it. Index structures that normalise keys away (cosine
similarity, MIPS reductions to angular search) therefore discard a signal
the model itself produced; in contrast, half-space range searching on the
raw key vectors preserves the magnitude and treats the score threshold $T$
in the units the attention layer uses. This is one of the reasons we model
sparse attention as range searching rather than as nearest-neighbour or
cosine retrieval.

\section{Index Design Challenges}\label{sec:challenges}

Following the reduction of Problem~\ref{problem:sparseattention}, our goal is to design an index over the KV cache that supports efficient halfspace range searching queries.
Classical data structures for range searching include geometric indexes such as KD-trees, range trees, partition trees~\cite{partitiontree}, and R-trees~\cite{rtrees}, which have been extensively studied in the literature (see the survey~\cite{rangesearching}). In contrast, production database systems primarily rely on B-tree variants, in particular B+-trees~\cite{btree}, which are widely used for range queries over ordered data.
However, these approaches are not well-suited to the LLM inference setting. KV cache indexing must operate under strict latency constraints during decoding, avoid introducing significant overhead, and be compatible with GPU-oriented execution. Moreover, the index must support dynamic updates, as new keys are generated and inserted at each decoding step.

We outline the key challenges of KV cache index design below.
Then, in the next section, we present our index, \textsc{Louver}, which addresses these challenges.

\paragraph{Challenge 1 (High Pruning Power with Full Recall).}
The primary goal of an index for halfspace range searching is to achieve sublinear query time by avoiding a full scan over all keys. A brute-force approach evaluates all $t-1$ keys in the KV cache and returns those satisfying the threshold $\tau$. To improve upon this, the index must be able to efficiently {\bf prune} large subsets of keys using cheap computations. 
At the same time, the index must guarantee full recall, i.e., zero false negatives. This requirement is critical in the LLM setting as discussed before. 
Therefore, the main challenge is to prune a large number of irrelevant keys while still retrieving all relevant ones. In addition, the pruning strategy should work well for the actual distribution of key and query vectors in transformer attention, so the index must be designed specifically for LLM inference.

\paragraph{Challenge 2 (Dynamic and Query-Adaptive).}
The index must be dynamic. During decoding, new keys are generated at every step, and in long-output tasks such as reasoning, most keys are produced on-the-fly during autoregressive generation rather than during prefilling. Therefore, the index must support efficient updates, allowing new keys to be inserted continuously without degrading performance.
In addition, the index must be query-adaptive. As observed earlier in Figure~\ref{fig:decoding}, the number of relevant keys varies across decoding steps and depends on the query. The range searching formulation naturally supports this by allowing arbitrary thresholds at query time. 

\paragraph{Challenge 3 (Efficiency and Integration).}
The final challenge is efficiency. The index should not introduce additional overhead during inference, especially in the decoding phase, which is already the main bottleneck. It should also avoid increasing the memory footprint of the KV cache, as memory usage is another critical constraint.
In addition, the index should make effective use of the underlying hardware, including both CPU and GPU (CUDA) environments, and help reduce memory bandwidth during attention computation—one of the main goals of sparse attention. It should also achieve speedups over highly optimized dense attention implementations such as FlashAttention~\cite{flashattention2}.
Finally, the index should be easy to integrate into existing LLM inference systems, requiring minimal changes and low implementation overhead.

In Appendix~\ref{app:ablation}, we present a general class of index structures with the potential for achieving strong pruning while maintaining full recall (addressing Challenge 1) and evaluate them through ablation studies on real LLM key and query distributions. The resulting insights guide the design of our final index, \textsc{Louver}, presented in the next section.
\section{Ablation on Index Design}
\label{app:ablation}

\subsection{High Pruning Power with Full Recall}

Achieving high pruning power while preserving zero false negatives requires the index to prune entire groups of keys cheaply without inspecting every key individually.
Inspired by classical geometric range-searching indexes~\cite{har2011geometric, rangesearching}, we define a general family of indexes parameterized by two choices: an \emph{enclosing geometry} and a \emph{grouping strategy}.

\paragraph{Enclosing.}
The key idea is to \emph{enclose} a subset of keys within a convex geometric shape $E$ (e.g., a bounding ball, an axis-aligned bounding box (AABB), or an ellipsoid).
At query time, given a halfspace $H(q,\tau) = \{k \in \mathbb{R}^d : \langle q, k\rangle \geq \tau\}$, we test whether $E$ intersects $H$.
If $E \cap H = \varnothing$, then no key inside $E$ can satisfy the threshold, and the entire group is pruned with zero false negatives.

\begin{definition}[Enclosing]\label{def:1}
An \emph{enclosing family} $\mathcal{E}$ is a collection of convex sets in $\mathbb{R}^d$ such that for any point set $P \subset \mathbb{R}^d$ and any $E \in \mathcal{E}$ with $P \subseteq E$:
\[
    E \cap H = \varnothing \;\Longrightarrow\; P \cap H = \varnothing.
\]
\end{definition}

This guarantee follows directly from convexity: if the enclosing shape does not intersect the halfspace, no point it contains can either.
The key practical requirement is that the intersection test $E \cap H \stackrel{?}{=} \varnothing$ be \emph{cheaper} than evaluating all $|P|$ dot products individually.
For a bounding ball $B(c, \rho)$, the test reduces to $\langle q, c\rangle + \rho\|q\|_2 < \tau$ — an $O(d)$ operation independent of the number of keys inside the ball (see Section~3 of the main paper for derivation).
For an AABB, the test requires two dot products ($g=2$) but often achieves tighter pruning than a ball.

\paragraph{Grouping.}
To apply enclosing, keys must first be partitioned into groups, each enclosed by a single shape.
A crucial design decision is the group size.

\begin{definition}[Grouping]\label{def:2}
A \emph{grouping} $\mathcal{G}(P)$ of a key set $P$ with $|P| = n$ is a partition of $P$ into $\lceil n/r \rceil$ disjoint subsets of size at most $r$, where $r = O(1)$ is a fixed group size constant.
\end{definition}

Using a fixed group size $r$ (rather than a fixed number of groups $K$) is essential.
With $K$ fixed groups and $n$ keys, each group contains $\Theta(n/K)$ keys; as $n$ grows, groups grow larger, their enclosing shapes become larger and more likely to intersect any query halfspace, and pruning power vanishes.
With fixed $r$, the number of groups grows as $\Theta(n/r)$, groups remain small, and enclosing shapes stay tight — preserving pruning power as the KV cache grows.
Note that prior work on KV cache clustering (e.g., for similarity search) typically uses a fixed $K$; our setting requires fixed $r$ instead.

\paragraph{Achieving high pruning.}
Given a grouping and an enclosing, the index filters groups by testing each enclosing against the query halfspace, then performs exact dot-product checks only on keys from surviving groups.
The asymptotic cost per query is proportional to $g/r + f_{\text{scan}}$, where $g$ is the gate cost per group (in dot-product equivalents) and $f_{\text{scan}}$ is the fraction of keys not pruned.
Speedup over brute-force is $1 / (g/r + f_{\text{scan}})$.
The empirical ablation below measures how different grouping strategies and enclosing choices affect $f_{\text{scan}}$ and speedup on real LLM key/query data.

\subsection{Empirical Ablation: Grouping Strategy and Subspace Count}
\label{app:ablation_table}

Table~\ref{tab:pruning_ablation_strategy} reports the fraction of keys scanned and the resulting speedup for four grouping strategies across $S \in \{2, 4, 8, 16\}$ subspaces, measured on Llama-3.2-3B-Instruct ($N \approx 4{,}500$, layer 15, $r{=}4$ keys per group).
All configurations achieve perfect recall ($=100\%$) by the halfspace enclosing guarantee.
Contiguous grouping assigns consecutive key positions to the same group, which concentrates positional locality and yields the strongest pruning.
At $S{=}16$, contiguous grouping scans only 16.3\% of keys — a $2.42\times$ speedup over brute-force — while interleaved achieves a comparable 17.2\% scan fraction ($2.37\times$).
PCA and random strategies achieve lower pruning at the same $S$, because they do not exploit the local key-space structure that contiguous grouping leverages.

\begin{table}[h]
\centering
\small
\caption{Pruning ablation: scanned fraction (\%) and speedup ($\times$) by grouping strategy and $S$. All entries: recall $= 100\%$, $r{=}4$, Llama-3.2-3B-Instruct, $N\approx4{,}500$.}
\label{tab:pruning_ablation_strategy}
\setlength{\tabcolsep}{6pt}
\begin{tabular}{lcccccccc}
\toprule
 & \multicolumn{2}{c}{$S=2$} & \multicolumn{2}{c}{$S=4$} & \multicolumn{2}{c}{$S=8$} & \multicolumn{2}{c}{$S=16$} \\
\cmidrule(lr){2-3}\cmidrule(lr){4-5}\cmidrule(lr){6-7}\cmidrule(lr){8-9}
\textbf{Strategy} & Scan\% & Speed & Scan\% & Speed & Scan\% & Speed & Scan\% & Speed \\
\midrule
Contiguous  & 89.7 & 0.87$\times$ & 72.0 & 1.03$\times$ & 46.1 & 1.41$\times$ & \textbf{16.3} & \textbf{2.42$\times$} \\
Interleaved & 90.9 & 0.86$\times$ & 80.1 & 0.95$\times$ & 55.0 & 1.25$\times$ & 17.2 & 2.37$\times$ \\
Random      & 90.9 & 0.86$\times$ & 82.7 & 0.93$\times$ & 62.9 & 1.14$\times$ & 23.8 & 2.05$\times$ \\
PCA         & 84.5 & 0.91$\times$ & 73.9 & 1.01$\times$ & 52.9 & 1.28$\times$ & 24.1 & 2.04$\times$ \\
\bottomrule
\end{tabular}
\end{table}

Table~\ref{tab:pruning_ablation_cluster} reports the effect of clustering method and enclosing geometry on scanned fraction, at $r{=}4$ keys per group ($N\approx8{,}500$, Llama-3.2-3B-Instruct, layer 14).
Among clustering methods, \texttt{kcenter} and \texttt{kmeans} produce tighter groups than \texttt{pq\_subspace}, yielding better pruning with ball and span-ball enclosings.
The axis-aligned bounding box (AABB) enclosing consistently provides the most aggressive pruning (39--61\% scanned fraction at $r{=}4$), at the cost of a higher gate evaluation cost ($g{=}2$ dot products per cluster vs.\ $g{=}1$ for ball and span-ball).
\textsc{Louver} adopts the ball enclosing with \texttt{subspace\_kcenter} clustering, which achieves a favorable balance between gate cost and pruning power.

\begin{table}[h]
\centering
\small
\caption{Ablation over clustering and enclosing geometry. Scanned fraction (\%) and recall, $r{=}4$, $N\approx8{,}500$, Llama-3.2-3B-Instruct, layer 14. Gate cost $g$: dot-product equivalents per cluster.}
\label{tab:pruning_ablation_cluster}
\setlength{\tabcolsep}{5pt}
\begin{tabular}{llccc}
\toprule
\textbf{Clustering} & \textbf{Enclosing} & \textbf{Scan (\%)} & \textbf{Recall} & \textbf{Gate $g$} \\
\midrule
\multirow{3}{*}{k-center}
  & Ball (centroid)  & 92.4 & 99.95\% & 1 \\
  & AABB             & 60.8 & 99.96\% & 2 \\
  & Span-ball        & 93.6 & 99.95\% & 1 \\
\midrule
\multirow{3}{*}{k-means}
  & Ball (centroid)  & 91.3 & 99.95\% & 1 \\
  & AABB             & 64.6 & 99.97\% & 2 \\
  & Span-ball        & 92.3 & 99.95\% & 1 \\
\midrule
\multirow{3}{*}{PQ subspace}
  & Ball (centroid)  & 95.7 & 99.81\% & 1 \\
  & AABB             & 65.7 & 99.83\% & 2 \\
  & Span-ball        & 89.3 & 99.74\% & 1 \\
\bottomrule
\end{tabular}
\end{table}

\section{How \textsc{Louver} Addresses the Design Challenges}
\label{sec:louver-analysis}
In this section, we discuss how our index, {\sc Louver}, addresses the design challenges introduced in Appendix~\ref{sec:challenges}.

\paragraph{Challenge 1 (High Pruning Power with Full Recall).}
\textsc{Louver} guarantees zero false negatives by construction.
For Query~1, pruning a cluster in \emph{any} subspace requires that no point inside its bounding ball can satisfy the per-subspace threshold, which in turn implies no point can satisfy the global threshold $\tau$ (as shown in Section~\ref{sec:query-processing}).
For Query~2, the TA stopping condition $U(d^*) < \tau$ certifies that every unseen key has full dot product below $\tau$.
In both cases the final exact dot-product check is exact, so the returned set $\mathcal{K}^*$ is precisely $\{k \in \mathcal{K} \mid \langle q, k \rangle \geq \tau\}$ with no omissions.

Pruning power comes from two complementary sources.
First, subspace decomposition amplifies pruning: a cluster must survive the ball test in \emph{all} $S$ subspaces to remain a candidate, so pruning in any single subspace eliminates the cluster entirely.
Crucially, each ball test operates in the reduced $d_s$-dimensional subspace rather than the full $d$-dimensional space.
In lower dimensions, a ball of fixed radius subtends a smaller solid angle relative to an arbitrary halfspace, so the probability of intersection is lower and more clusters are pruned per subspace.
Second, balanced PCA tree grouping ensures that clusters are compact --- the dominant variance axis aligns nearby keys --- so bounding balls are tight and the ball-halfspace test rejects many clusters cheaply.
Together these properties yield strong pruning on real LLM key distributions, as confirmed empirically in Appendix~\ref{app:ablation}.

\paragraph{Challenge 2 (Dynamic and Query-Adaptive).}
\textsc{Louver} handles the dynamic nature of decoding through its buffer-and-update mechanism (Section~\ref{sec:dynamic-updates}).
New keys are appended to buffer $\mathcal{B}$ and attended densely with no filtering, ensuring correctness from the moment a key is generated.
Once the buffer fills, the same index construction procedure runs on the buffered keys, and the resulting groups are appended directly to the main index without any rebuild.

Query-adaptivity is inherent to the range searching formulation: both query algorithms accept a threshold value $\tau$ at runtime, with no assumptions on the number of relevant keys.
The retrieved set $\mathcal{K}^*$ adapts automatically to the query and the current context, handling the oscillating score distributions observed in Figure~\ref{fig:decoding} without any modification to the index.

\paragraph{Challenge 3 (Efficiency and Integration).}
\textsc{Louver} is designed for hardware efficiency at every level.
Fixed group size $r$ enables fully regular memory access patterns during both the ball filter and the exact check, avoiding conditional branching and enabling vectorized execution.
On GPU, the index updates run concurrently with attention, hiding their cost behind the main computation.
The ball filter operates only on the $K = n/r$ group centers, which is a factor of $r$ smaller than the full key array; the full keys are accessed only for the surviving candidate set, directly reducing memory bandwidth pressure.
On CPU, the amortized update cost is $O(S \cdot d_s / r)$ per step — constant in $n$ — while query cost over the index is sublinear in $n$ for typical LLM score distributions.

All query and update operations are backed by hardware-optimized CUDA and CPU kernels.
As shown in Section~\ref{sec:experiments}, {\em \textsc{Louver} outperforms even highly optimized dense attention implementations such as FlashAttention~\cite{flashattention2}}.

\section{Pseudo-codes}
\label{app:pseudocodes}

\begin{algorithm}[H]
\caption{\textsc{Louver} Index Construction}
\label{alg:build}
\begin{algorithmic}[1]
\Require Keys $\mathcal{K} = \{k_1,\ldots,k_n\} \subset \mathbb{R}^d$,
         subspace count $S$, group size $r$
\Ensure Index $\mathcal{I} = \bigl(\{c_{s,i},\, \rho_{s,i},\, a_{s,\cdot}\}_{s=1}^{S}\bigr)$
\State Compute slices $(\delta_0, \ldots, \delta_S)$ with $\delta_s - \delta_{s-1} \in \{\lfloor d/S \rfloor, \lceil d/S \rceil\}$
\For{$s = 1$ \textbf{to} $S$}
    \State $k_j^{(s)} \leftarrow \pi_s(k_j)$ for all $j$ \Comment{project onto subspace $s$}
    \State $\{a_{s,j}\} \leftarrow \textsc{BalancedPCATree}\!\bigl(\{k_j^{(s)}\},\, r\bigr)$ \Comment{group assignments}
    \For{$i = 1$ \textbf{to} $K$}
        \State $c_{s,i} \leftarrow \frac{1}{r}\sum_{j:\, a_{s,j}=i} k_j^{(s)}$ \Comment{group center}
        \State $\rho_{s,i} \leftarrow \max_{j:\, a_{s,j}=i} \| k_j^{(s)} - c_{s,i} \|_2$ \Comment{ball radius}
    \EndFor
\EndFor
\State \Return $\mathcal{I}$
\end{algorithmic}
\end{algorithm}

\begin{algorithm}[H]
\caption{\textsc{BalancedPCATree}}
\label{alg:pca-tree}
\begin{algorithmic}[1]
\Require Points $P = \{p_1,\ldots,p_m\} \subset \mathbb{R}^{d_s}$,
         group size $r$,
         next cluster id $\mathit{offset}$ (default $0$),
         assignment array $a[\,]$ (shared, modified in place)
\Ensure $a[j]$ set for all $j \in \{1,\ldots,m\}$
\If{$m \leq r$} \Comment{leaf: assign all points to one group}
    \State $a[j] \leftarrow \mathit{offset}$ for all $j \in \{1,\ldots,m\}$
    \State \Return
\EndIf
\State $\ell \leftarrow \lfloor m/2 \rfloor$ \Comment{left subtree size}
\State $\hat{e} \leftarrow \argmax_{e \in \{e_1,\ldots,e_{d_s}\}} \operatorname{Var}_{p \in P}(\langle p, e \rangle)$
    \Comment{dominant coordinate axis}
\State Sort $P$ by $\langle p, \hat{e} \rangle$; let $L \leftarrow P_{1:\ell}$, $R \leftarrow P_{\ell+1:m}$
\State \textsc{BalancedPCATree}$(L,\; r,\; \mathit{offset},\; a)$
\State \textsc{BalancedPCATree}$(R,\; r,\; \mathit{offset} + \lceil \ell / r \rceil,\; a)$
\end{algorithmic}
\end{algorithm}

\begin{algorithm}[H]
\caption{\textsc{Louver} Query 1 — Full-Subspace Ball Filter}
\label{alg:query-full}
\begin{algorithmic}[1]
\Require Query $q \in \mathbb{R}^d$, per-subspace thresholds $(\tau_s)_{s=1}^S$,
         global threshold $\tau$, index $\mathcal{I}$, buffer $\mathcal{B}$, scale $\sigma$
\Ensure Attention output $o \in \mathbb{R}^d$
\For{$s = 1$ \textbf{to} $S$} \Comment{ball filter per subspace}
    \State $\mathcal{L}_s \leftarrow \bigcup \bigl\{\mathcal{C}_{s,i} : \langle \pi_s(q), c_{s,i} \rangle + \rho_{s,i}\|\pi_s(q)\|_2 \geq \tau_s \bigr\}$
\EndFor
\State $\mathcal{L} \leftarrow \mathcal{L}_1 \cap \mathcal{L}_2 \cap \cdots \cap \mathcal{L}_S$ \Comment{AND across subspaces}
\State $\mathcal{L}^* \leftarrow \{j \in \mathcal{L} : \langle q, k_j \rangle \geq \tau\}$ \Comment{exact dot-product check}
\State $o \leftarrow \textsc{Attention}(q,\; \mathcal{K}[\mathcal{L}^*] \cup \mathcal{B},\; \sigma)$
\State \Return $o$
\end{algorithmic}
\end{algorithm}

\begin{algorithm}[H]
\caption{\textsc{Louver} Query 2 — TA Filter}
\label{alg:query-ta}
\begin{algorithmic}[1]
\Require Query $q \in \mathbb{R}^d$, threshold $\tau \in \mathbb{R}$,
         index $\mathcal{I}$, buffer $\mathcal{B}$, scale $\sigma$
\Ensure Attention output $o \in \mathbb{R}^d$
\For{$s = 1$ \textbf{to} $S$}
    \State $f_{s,i} \leftarrow \langle \pi_s(q), c_{s,i} \rangle + \rho_{s,i}\,\|\pi_s(q)\|_2$ \quad for all $i$
    \State $\sigma_s \leftarrow \operatorname{argsort}_i(-f_{s,i})$ \Comment{descending upper bounds}
\EndFor
\State $\mathcal{L} \leftarrow \emptyset$,\quad $d \leftarrow 0$
\Repeat
    \State $d \leftarrow d + 1$
    \For{$s = 1$ \textbf{to} $S$}
        \State $\mathcal{L} \leftarrow \mathcal{L} \cup \mathcal{C}_{s,\,\sigma_s(d)}$ \Comment{add all members of scanned cluster}
    \EndFor
    \State $U \leftarrow \sum_{s=1}^S f_{s,\,\sigma_s(d)}$ \Comment{upper bound on unseen keys}
\Until{$U < \tau$}
\State $\mathcal{L}^* \leftarrow \{j \in \mathcal{L} : \langle q, k_j \rangle \geq \tau\}$ \Comment{exact dot-product check}
\State $o \leftarrow \textsc{Attention}(q,\; \mathcal{K}[\mathcal{L}^*] \cup \mathcal{B},\; \sigma)$
\State \Return $o$
\end{algorithmic}
\end{algorithm}

\section{Update Process}
\label{app:update}

\subsection{GPU Update}

On GPU, the incremental update runs concurrently with the ongoing attention computation using a two-stream design.
The index maintains a pre-allocated arena with sufficient capacity for all expected update rounds; no memory reallocation occurs during decoding.

The update is split into two phases:
\begin{itemize}[leftmargin=*, nosep]
    \item \textbf{Phase 1} (side stream, concurrent with attention): the $B$ buffered keys are clustered into $B/r$ new groups; their centers, radii, assignments, and key/value data are written into the unused tail of the arena.
    New slots are marked \emph{invalid} throughout Phase~1, so the attention kernel on the main stream treats them as empty and skips them — no locking or synchronization is required.
    \item \textbf{Phase 2} (publish, on attention stream): after the attention stream waits on the update-done event, the invalid flags for the new slots are flipped atomically and the live-count counters are advanced.
    The new groups become visible to all subsequent queries from this point on.
\end{itemize}
Because Phase~1 and attention run concurrently, the wall-clock cost of an update round is hidden behind the attention computation in the common case.

\subsection{CPU Update}

On CPU, the update runs synchronously after every $B$ decoding steps.
The total work per update round is $O(S \cdot B \cdot \log(B/r))$ for the balanced PCA tree builds across all $S$ subspaces, plus $O(S \cdot B \cdot d_s)$ for computing centers and radii.
Amortized over the $B$ steps between triggers, the per-step cost is $O(S \cdot d_s \cdot \log(B/r))$, which is constant in $n$ and negligible relative to the $O(n \cdot d)$ cost of dense attention.

\section{Kernel Implementation Details}
\label{app:kernels}

All performance-critical operations in \textsc{Louver} are implemented as custom CUDA and CPU kernels.
On GPU, the TA filter runs as a three-stage fused pipeline (score, depth, alive compaction) in a single cooperative kernel launch, followed by a sparse FlashAttention-style kernel over the surviving candidate set.
On CPU, the filter and attention are implemented in C++ with AVX/FP16 intrinsics.

\subsection{GPU Filter Pipeline}

The TA filter is implemented as three CUDA kernels fused into a single cooperative kernel launch via \texttt{cudaLaunchCooperativeKernel}, with \texttt{cooperative\_groups::grid\_group::sync()} barriers separating phases.
This avoids separate kernel dispatch overhead and keeps intermediate data in shared memory across phases.

\paragraph{Score kernel (Phase 1).}
For each query head $h_q$ and subspace $s$, this kernel computes $f_{s,i} = \langle \pi_s(q), c_{s,i} \rangle + \rho_{s,i}\|\pi_s(q)\|_2$ for all $K$ clusters and retains the top-$L$ scores.
Each thread maintains a min-heap of size $L/\text{IPT\_L}$ over its assigned clusters; a CUB BlockRadixSort then produces the globally sorted top-$L$ per $(h_q, s)$.
This avoids sorting all $K$ clusters — only $L \ll K$ entries enter the sort.

\textbf{GQA reuse.}
The grid is launched as $(H_{kv}, 4)$ rather than $(H_q, 4)$.
Each block streams the $K$ centers for one $(h_{kv}, s)$ pair from global memory \emph{once} and scores all $G$ query heads that share this KV head simultaneously.
This reduces center GMEM traffic by a factor of $G$ (approximately $3\times$ for Llama, $7\times$ for Qwen).

\paragraph{Depth kernel (Phase 2).}
One block per query head scans the sorted top-$L$ scores to find the TA stopping depth $d^*$ (smallest $d$ with $U(d) < \tau$) and writes a per-subspace bitmask \texttt{parent\_alive\_bitmap[}$H_q$\texttt{, 4, K\_words]} marking which clusters survive.
This also resets the live-count counter used by Phase 3.

\paragraph{Alive compaction kernel (Phase 3).}
Grid: $(H_q, \text{N\_TILES})$ with TILE\_N\,=\,2048 keys per tile.
Each thread loads packed cluster assignments using 128-bit vectorized loads (4 assignments per \texttt{ld.b128}, folding the invalid-key sentinel into the packing).
The bitmask is read once from global memory into shared memory per tile; per-key lookups are then shared-memory reads.
Alive keys within each warp are compacted via \texttt{\_\_ballot\_sync}; a CUB BlockScan computes prefix offsets; each lane writes its alive key directly to the output \texttt{live\_idx} array in global memory.
One \texttt{atomicAdd} per block (not per warp) claims a contiguous output range.

\subsection{GPU Sparse Attention Kernel}

The sparse attention kernel receives the compact \texttt{live\_idx} list from the filter and computes scaled dot-product attention restricted to the surviving keys.
The sequence is split into $P$ splits; each split computes partial softmax statistics $(m_p, \ell_p, o_p)$ in the FlashAttention~\cite{flashattention2} reduction scheme.
A final combine pass merges the $P$ partials into the output $o$.
Buffer keys $\mathcal{B}$ are appended to each split to ensure they are always attended densely.

\subsection{CPU Kernels}

The CPU filter is implemented in C++ with AVX2/FP16 intrinsics.
Ball scores are computed with vectorized dot products over the $d_s$-dimensional center vectors; the AND-across-subspaces pruning is done with 64-bit bitmask operations.
The CPU attention kernel performs a fused dot-product + softmax over the surviving candidates, with inner loops fully unrolled at compile time using the fixed group size $r$ as a template parameter.
Both kernels operate entirely within L2/L3 cache for typical sequence lengths, since only the $K = n/r$ centers need to be streamed during filtering.

\section{Discussion}
\label{app:limit}

\paragraph{Integration with inference engines.}
\textsc{Louver} is designed to integrate with existing LLM inference systems with minimal changes.
vLLM~\cite{vllm} manages the KV cache in fixed-size pages; by reordering children within each parent group into contiguous memory, child fetches from GPU parents remain cache-friendly and hardware-efficient — directly compatible with vLLM's block allocator.
We also provide a drop-in cache and attention module compatible with HuggingFace Transformers and PyTorch, enabling use with any LLM architecture without modification.

\paragraph{Broader impact.}
This work identifies false negatives in sparse attention as a first-class source of accuracy degradation, a direction not well-studied in prior work.
We hope the halfspace range-searching framing opens a principled path for future work on sparse attention correctness, beyond heuristic top-$k$ or eviction strategies.

\paragraph{Limitations.}
\textsc{Louver}'s primary design objective is retrieval correctness and low latency; reducing GPU memory usage is a secondary goal.
The index introduces parent cluster centers as additional GPU-resident data: a memory overhead that is small relative to the full KV cache (28\% at 1\,GB; see Table~\ref{tab:offload}), but non-zero.

One approach to eliminate this overhead entirely is to replace synthetic cluster centers with actual keys: select the key closest to each group's centroid as the parent representative and store only a reference (index) to it rather than a new $d$-dimensional vector.
This is analogous to how hierarchical ANN indexes (e.g., HNSW~\cite{hnsw}) use existing data points as navigating nodes, avoiding any additional storage.


\end{document}